\documentclass[10pt,twocolumn,letterpaper]{article}

\PassOptionsToPackage{dvipsnames,table}{xcolor}
\usepackage[pagenumbers]{cvpr} %

\usepackage{amsfonts}
\usepackage{amsmath}
\usepackage{tabularx}
\usepackage{multirow}
\usepackage{microtype}
\usepackage{nicefrac}
\usepackage{pifont}
\usepackage{comment}
\usepackage{soul}
\usepackage{cuted}            %
\usepackage{tikz}
\usetikzlibrary{spy}
\usepackage{makecell}
\usepackage{adjustbox}
\usepackage{booktabs}
\usepackage{pgf}
\usepackage{wrapfig}
\usepackage[most]{tcolorbox}
\usepackage{listings}
\definecolor{green}{RGB}{0,150,10}

\newcommand{\cmark}{\textcolor{blue}{\ding{51}}}%
\newcommand{\xmark}{\textcolor{red}{\ding{55}}}%

\newcommand{\figLabel}{Figure~}

\newcommand{\secLabel}{Section~}
\newcommand{\tblLabel}{Table~}
\newcommand{\mysection}[1]{\noindent\textbf{#1.}}
\newcommand{\supp}{\textit{supplementary material}\xspace}

\newlength\mytmplen

\definecolor{GTAmber}{RGB}{214,133,0}  %

\usepackage{xspace}
\definecolor{mri}{RGB}{66,133,244}
\definecolor{ct}{RGB}{220,53,69}
\definecolor{pet}{RGB}{255,140,0}
\newcommand{\MRI}{\textcolor{mri}{\textbf{\texttt{MRI}}}}
\newcommand{\CT}{\textcolor{ct}{\textbf{\texttt{CT}}}}
\newcommand{\PET}{\textcolor{pet}{\textbf{\texttt{PET}}}}

\definecolor{goodred}{HTML}{e67c73}
\definecolor{goodyellow}{HTML}{ffd666}
\definecolor{goodgreen}{HTML}{57bb8a}

\newcommand{\cc}[2]{%
  \pgfmathsetmacro{\ccp}{max(0,min(100,100*(#1)/(#2)))}%
  \ifdim\ccp pt<50pt
    \pgfmathsetmacro{\cclow}{2*\ccp}\xdef\cclow{\cclow}%
    \cellcolor{goodyellow!\cclow!goodred}{#1}%
  \else
    \pgfmathsetmacro{\cchigh}{2*(\ccp-50)}\xdef\cchigh{\cchigh}%
    \cellcolor{goodgreen!\cchigh!goodyellow}{#1}%
  \fi}
\newcommand{\ccb}[2]{%
  \pgfmathsetmacro{\ccp}{max(0,min(100,100*(#1)/(#2)))}%
  \ifdim\ccp pt<50pt
    \pgfmathsetmacro{\cclow}{2*\ccp}\xdef\cclow{\cclow}%
    \cellcolor{goodyellow!\cclow!goodred}{\textbf{#1}}%
  \else
    \pgfmathsetmacro{\cchigh}{2*(\ccp-50)}\xdef\cchigh{\cchigh}%
    \cellcolor{goodgreen!\cchigh!goodyellow}{\textbf{#1}}%
  \fi}
\newcommand{\oursrow}{\rowcolor[HTML]{EFEFEF}}



\renewcommand{\supp}{appendix\xspace}   %

\definecolor{cvprblue}{rgb}{0.21,0.49,0.74}
\usepackage[breaklinks,colorlinks,allcolors=cvprblue]{hyperref}   %


\newcommand{\ours}{RadContour\xspace}           %
\newcommand{\haiku}{Claude Haiku~5.5\xspace}    %
\newcommand{\sonnet}{Claude Sonnet~5.5\xspace}  %
\newcommand{\fable}{Claude Fable~5.2\xspace}    %

\newcommand{\flash}{Gemini~3.8 Flash\xspace}
\newcommand{\oursh}{\ours{} (Haiku~5.5)\xspace}     %
\newcommand{\ourss}{\ours{} (Sonnet~5.5)\xspace}
\newcommand{\oursf}{\ours{} (Fable~5.2)\xspace}   %
\newcommand{\second}[1]{\underline{#1}}
\newcommand{\eqn}[1]{Eq.~(\ref{#1})}
\newcommand{\na}{{--}}

\newcommand{\sd}[1]{{\scriptsize$\pm$#1}}

\newlength{\qgw}

\newcommand{\qimh}[1]{\includegraphics[width=\qgw,height=\qgw,keepaspectratio]{imgs/#1}}
\newcommand{\qhd}[1]{{\scriptsize\textbf{#1}}}
\newcommand{\qrow}[2]{\rotatebox{90}{\parbox{\qgw}{\centering\scriptsize\textbf{#1}\\[-1pt]{\tiny\emph{``#2''}}}}}

\title{RadContour: Can Frozen Multimodal LLMs Segment 3D Medical Images?}

\author{Abdullah Hamdi \qquad Carlos Hinojosa\\
King Abdullah University of Science and Technology (KAUST)
}

\begin{document}
\maketitle

\let\arxivendabstract\endabstract
\def\endabstract{\unskip\space The code is available at
  \href{https://abdullahamdi.com/radcontour/}{https://abdullahamdi.com/radcontour/}.\arxivendabstract}
\begin{abstract}
Segmenting anatomies and pathologies in 3D medical images underpins diagnosis, treatment planning, and disease
monitoring. Yet dense 3D labels are among the scarcest data in medicine, and segmentation models trained on them generalize poorly
beyond their training distribution. General-purpose Multimodal Large Language Models (MLLMs) are not trained to segment volumes, but
they already know much of the anatomy and radiology a clinician describes in words. We present \ours, a training-free
pipeline that turns a frozen MLLM into a text-prompted 3D segmenter of CT, MRI, and PET volumes, drawing
outlines on slices, much as radiologists contour a lesion. The MLLM is
the \emph{only} vision component. It plans the target from a free-text prompt, localizes it in 3D from coarse to fine,
and traces closed outlines on a few keyframes. Deterministic geometry then assembles the outlines into a 3D mask.
We evaluate \ours zero-shot on 17
public CT, MRI, and PET benchmarks spanning 923 volumes. On nine tumor and lesion benchmarks, \ours matches the mean
Dice of the trained state of the art on identical cases (58.7 vs.\ 58.4) and more than doubles that of every other
trained text-promptable model. On soft-tissue sarcoma, it nearly doubles the Dice of the best trained model (76.5 vs.\
39.3). Across 25 backbone MLLMs, accuracy follows backbone strength with an unchanged pipeline.
\end{abstract}

\let\endabstract\arxivendabstract

\section{Introduction} \label{sec:introduction}
Segmenting organs and lesions in CT, MRI, and PET volumes underlies diagnosis, radiotherapy planning, and follow-up.
Unlike natural images, medical volumes with dense voxel labels are scarce. They are expensive to acquire, protected by
privacy regulations, and laborious for experts to delineate slice by slice~\cite{ronneberger2015unet,nnUNet,Bourigault_2025_ICCV}.
The field's answer has been to curate ever-larger labeled collections and train ``foundation'' segmenters on
them~\cite{wasserthal2023totalsegmentator,medsam,MedSAM2,segvol,SAT,biomedparse,rokuss2025voxtell}. The state of
the art in text-prompted 3D segmentation, VoxTell~\cite{rokuss2025voxtell}, is trained on more than 62K volumes from
158 datasets on 64 GPUs for six days. Yet these models remain bound to their label sets and training
distributions. On concepts that are absent or defined differently during training, such as soft-tissue sarcoma or bladder
carcinoma, even the strongest model reaches only about 40 and 25 Dice, respectively. Earlier text-promptable models fall to near zero.
Every new structure, modality, or labeling convention restarts the cycle of acquisition, annotation, and training.

\begin{figure}[t]
  \centering
  \IfFileExists{imgs/teaser/teaser.pdf}{%
    \includegraphics[width=\linewidth]{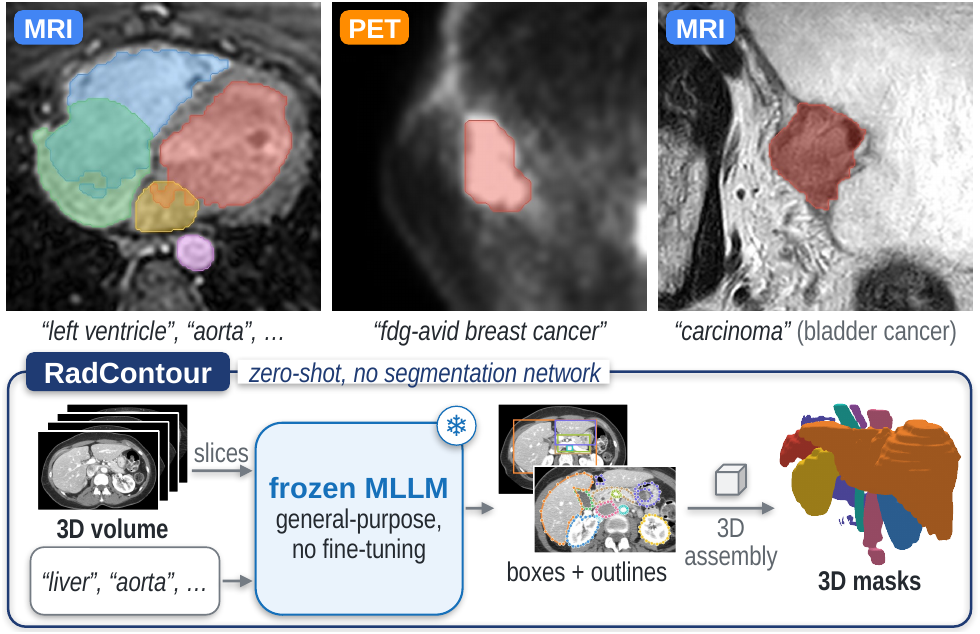}}{%
    \includegraphics[width=0.6\linewidth]{example-image}}
  \vspace{-14pt}
  \caption{\small\textbf{3D medical segmentation from text with a frozen, general-purpose MLLM.} \ours turns an MLLM that was not built for 3D segmentation into an accurate zero-shot, text-prompted 3D segmenter of CT, MRI, and PET
  volumes. The MLLM reads rendered slices and returns boxes and point outlines, which deterministic geometry assembles
  into a 3D mask.}
  \label{fig:teaser}
  \vspace{-8pt}
\end{figure}

General-purpose multimodal large language models (MLLMs)~\cite{GPT4,comanici2025gemini,anthropic2026claude55} are not built for medical
segmentation. They are, however, trained on a breadth of text and images that no medical dataset can match, and they
already encode much of the anatomy and radiology a clinician would describe in words~\cite{HamdiAbd_ColonBench_MICCAI2026}. Until recently, they were assumed
to need a trained mask decoder to produce dense output~\cite{lai2024lisa,rasheed2024glamm}.
LLMSeg~\cite{hamdi2026llmseg} showed that this is not necessary in 2D. A frozen MLLM can segment natural images by
emitting box-confined B-spline outlines that deterministic code rasterizes. This opens a different path for medicine
(\figLabel\ref{fig:teaser}).
The medical knowledge of such a segmenter comes free from a general model, and its accuracy grows as the
backbone improves. Its cost is a per-volume inference bill, not an up-front data and training budget, and that
bill falls as MLLM prices drop~\cite{cottier2025llmprices}.

\begin{figure}[t]
  \centering
  \includegraphics[width=\linewidth]{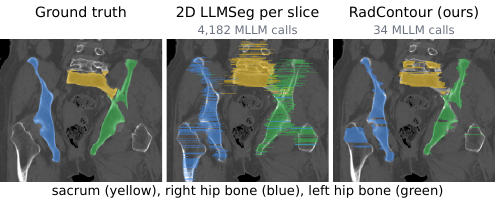}
  \vspace{-14pt}
  \caption{\small\textbf{Per-slice MLLM segmentation of a volume and its failure modes.} We show a frontal cut
  through a pelvic CT volume. A 2D MLLM segmenter (LLMSeg~\cite{hamdi2026llmseg}), applied independently to every axial slice and
  stacked, marks slices that do not contain the target, labels the femoral heads as hip bone, and is inconsistent between adjacent slices.
  \ours, with the same MLLM, localizes the target in 3D before accurately tracing its boundary and needs over 100 times fewer MLLM calls.}
  \label{fig:slice}
  \vspace{-8pt}
\end{figure}

Does this 2D ability carry over to volumes? The direct route is to run the 2D segmenter on every slice and stack the
masks. We find that this fails (\figLabel\ref{fig:slice}). On ten pelvic CT volumes, LLMSeg applied to every axial
slice reaches 54.8 Dice and needs 4{,}182 MLLM calls per volume. It finds the bones, but it also marks slices that do
not contain them, mistakes neighboring bones for the target, and produces outlines that jump from one slice to the next. A volume
is not a stack of independent images. Most of its slices do not show the target at all. The model must decide where
the target begins and ends, recognize it where it looks nothing like a textbook drawing, and stay consistent across
slices. It must do so in calls that each hold only a few images.

We present \ours (\figLabel\ref{fig:teaser}), a training-free pipeline that turns a frozen MLLM into a
text-prompted 3D segmenter. The MLLM is the only vision component, and it works on the volume in three steps. It first
writes a short plan of the target from the prompt. It then localizes the target in 3D, coarsely on a sparse set of
views and finely on dense slices. Only then does it trace the outline on a few keyframes, each shown as a full slice
to recognize the target and a zoomed crop to draw its boundary. Deterministic geometry interpolates the keyframe outlines
through signed distance functions into a 3D mask and snaps its boundary to the image intensities. Nothing is trained.
This mirrors radiological contouring, the way clinicians delineate a lesion or organ by hand. They scroll through the
volume, check the target in other views, and draw its boundary on one slice after another. \ours segments by drawing
outlines in the same way, which sets it apart from segmentation networks that label every voxel in one forward pass.
The only interface between the MLLM and the volume consists of rendered images going in and boxes or point lists coming out.
To the best of our knowledge, we show for the first time that the text output of an MLLM can produce
high-quality 3D segmentation masks of anatomies and pathologies in medical volumes.
Using the same MLLM as the slice-by-slice baseline, \ours reaches 82.5 Dice on the same ten volumes in 34 calls.

\noindent\textbf{Contributions.}
\textbf{(i)} We propose \ours, the first training-free pipeline that segments CT, MRI, and PET volumes from free text
with a frozen, general-purpose MLLM as its only vision component.
\textbf{(ii)} Evaluated zero-shot on 17 public benchmarks, \ours matches the mean Dice of the trained state of the art
on nine tumor and lesion benchmarks (58.7 vs.\ 58.4 on identical cases). On soft-tissue sarcoma, it nearly doubles that score
(76.5 vs.\ 39.3).
\textbf{(iii)} We analyze every stage of the pipeline thoroughly. Across 25 backbone MLLMs, accuracy follows the strength of the backbone with no change to the pipeline.

\section{Related Work} \label{sec:related}

\begin{figure*}[t]
  \centering
  \IfFileExists{imgs/pipeline/pipeline.pdf}{%
    \includegraphics[width=\textwidth]{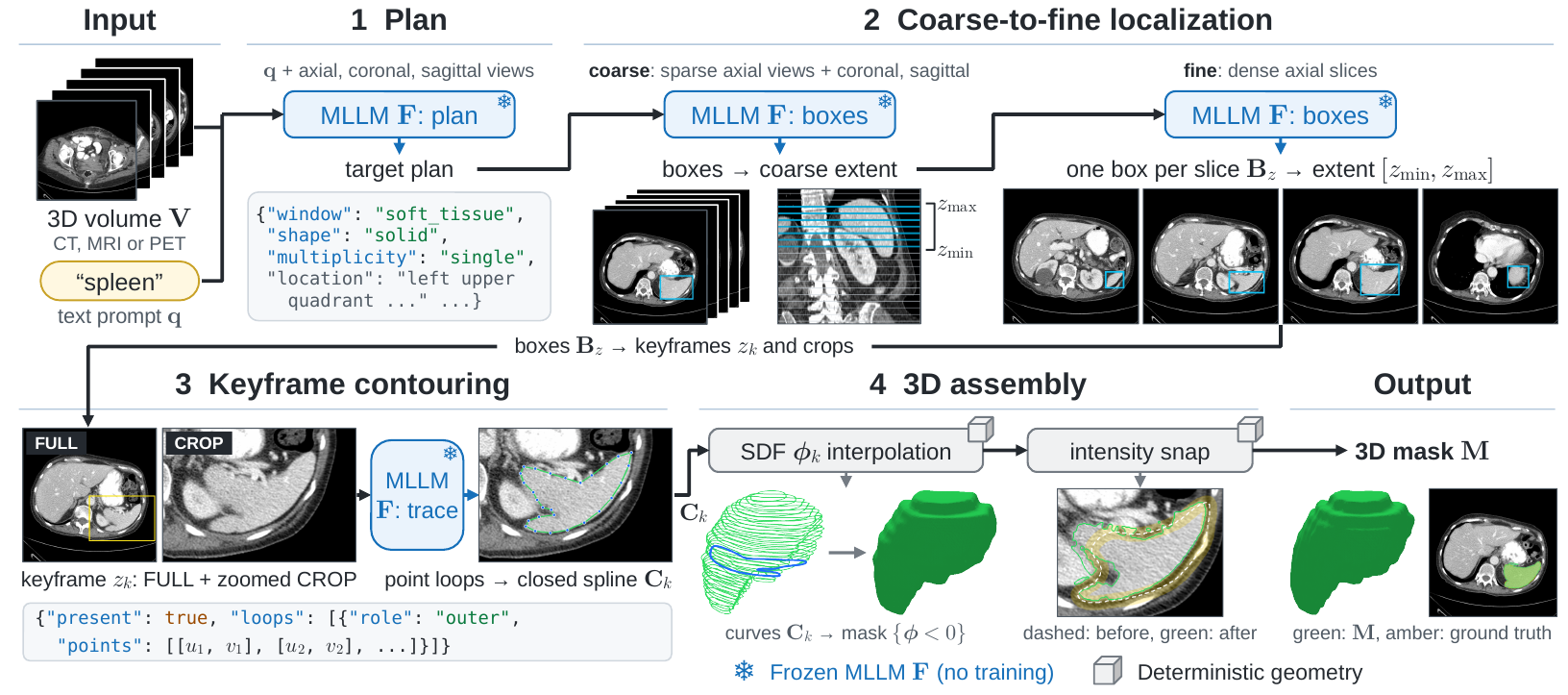}}{%
    \includegraphics[width=0.5\textwidth]{example-image}}
  \caption{\small\textbf{Overview of \ours.} From a text prompt $\mathbf{q}$ and a volume $\mathbf{V}$, a frozen MLLM $\mathbf{F}$ first plans the target. Next, it localizes the target coarsely on sparse views and finely on dense slices to fix the
  extent $[z_{\min},z_{\max}]$. It then traces closed outlines on keyframes $z_k$, each shown as a full slice plus a zoomed crop.
  Deterministic geometry code interpolates the keyframe curves $\mathbf{C}_k$ through signed distances $\boldsymbol{\phi}_k$ into a 3D mask
  and snaps its boundary to the image, giving $\mathbf{M}$. The steps follow how radiologists contour a lesion by hand.
  They find it while scrolling through the volume, check it in other views, and draw its boundary on selected slices.
  Thumbnails are real intermediate outputs of one case.}
  \label{fig:pipeline}
\end{figure*}

\mysection{3D medical image segmentation and its data bottleneck}
Volumetric segmentation has been dominated by supervised encoder-decoders since U-Net~\cite{ronneberger2015unet}. The
self-configuring nnU-Net~\cite{nnUNet,nnunet_revisited} and transformer hybrids such as
Swin-UNETR~\cite{Hatamizadeh2022SwinUS} reach expert-level accuracy when trained and tested on the same distribution.
However, their generalization is bounded by labeled training data, which is scarce in 3D. Collections such as
TotalSegmentator~\cite{wasserthal2023totalsegmentator}, AbdomenAtlas~\cite{abdomenatlas8k,AbdomenAtlas}, and
UKBOB~\cite{Bourigault_2025_ICCV} scale labels through model-in-the-loop annotation, yet they still cover a fixed set
of structures in a few modalities. Promptable ``segment anything'' models~\cite{kirillov2023segment,ravi2024sam2segmentimages}
were adapted to medicine at scale in MedSAM~\cite{medsam}, MedSAM2~\cite{MedSAM2,zhu2024medical}, SAM-Med3D~\cite{sammed3d},
VISTA3D~\cite{vista3d}, and nnInteractive~\cite{nninteractive}. Yet they still need a spatial prompt (point, box, or scribble)
for every target. \ours removes both requirements. It trains no segmentation network on voxel labels, and its only
prompt is text. The MLLM localizes the target itself, so no user supplies a point or box. Its outlines replace a
learned mask decoder.

\mysection{Text-promptable medical segmentation}
Text was first used to select among fixed output heads~\cite{liu2023clip,CAT,multitalent}. It then became a free-form
prompt fused with image features, in 2D by BiomedParse~\cite{biomedparse} and BoltzFormer~\cite{boltzformer}, and in 3D
by SegVol~\cite{segvol}, Text3DSAM~\cite{xin2025text3dsam}, and SAT~\cite{SAT}. VoxTell~\cite{rokuss2025voxtell} fuses
text at every decoder stage, trains on 62K volumes, and sets the state of the art on held-out benchmarks. We follow its
evaluation protocol. All of these models learn the link between words and voxels from paired medical data, so their
vocabulary and robustness are tied to their training corpus. \ours trains nothing. The link comes from a
general-purpose MLLM, and the only medical data the pipeline uses is the test volume.

\mysection{MLLMs for dense prediction}
MLLMs ground and describe both natural~\cite{GPT4,comanici2025gemini,bai2025qwen3} and medical
images~\cite{li2023llavamed,tu2024medpalmm}. Medical MLLMs for 3D volumes (RadFM~\cite{wu2023radfm},
M3D-LaMed~\cite{bai2024m3d}, CT-CHAT~\cite{hamamci2024ctchat}) and medically tuned open models
(MedGemma~\cite{sellergren2025medgemma}) are trained on medical image-text pairs for report generation and question
answering. When they support segmentation, a trained 3D decoder produces the mask. LLMSeg~\cite{hamdi2026llmseg} removed
the decoder in 2D. There, a frozen MLLM detects the object, then traces and refines box-confined B-spline outlines that
deterministic code rasterizes. Run on every slice of a volume, this recipe fails
(\figLabel\ref{fig:slice}). \ours builds on the idea and adds what 3D requires. It plans the target from text and
fixes its 3D extent by coarse-to-fine localization over slices. It then traces only a few keyframes per call, each shown as a paired full slice and zoomed
crop. Finally, it replaces 2D rasterization with 3D signed-distance assembly and
intensity snapping.

\section{Method} \label{sec:method}

\ours maps a 3D volume $\mathbf{V}:\mathbb{R}^3\to\mathbb{R}$ (CT, MRI, or PET) and a free-text prompt $\mathbf{q}$ to a
binary mask $\mathbf{M}:\mathbb{R}^3\to\{0,1\}$ (\figLabel\ref{fig:pipeline}). Here, $z$ is the position of an axial
slice along the body axis, and $\mathbf{u}=(u,v)$ is a pixel within the slice, so $\mathbf{V}(\mathbf{u},z)$ is one
voxel. A frozen MLLM $\mathbf{F}$ is the only vision component, and it never sees raw voxels or features. Each
call receives rendered 2D views of $\mathbf{V}$, captioned with their positions, and returns a JSON object with boxes or
point lists in view pixels that are mapped back to $(\mathbf{u},z)$.
The design answers the failure of slice-by-slice segmentation (\figLabel\ref{fig:slice}). The MLLM first
decides what to look for and where the target lies in 3D. Only then does it trace on a few keyframes instead of every
slice, while geometry, not $\mathbf{F}$, keeps neighboring slices consistent. This follows how radiologists contour a
lesion. They find it while scrolling through the volume, check it in coronal and sagittal views, and draw its boundary
on selected slices, leaving the slices in between to interpolation.

\begin{table*}[t]
\centering
\setlength{\tabcolsep}{4.2pt}
\begin{adjustbox}{max width=\textwidth}
\begin{tabular}{lc|cc|cc|ccccc|c}
\toprule
Method & \shortstack{3D Seg.\\Network} & \shortstack{Bladder\\Cancer} & \shortstack{Esoph.\\Tumor} & \shortstack{Breast\\Cancer} & Sarcoma & \shortstack{Lung\\Tumor} & \shortstack{Multiple\\Sclerosis} & \shortstack{Adrenal\\Tumor} & \shortstack{Liver\\Lesions} & \shortstack{Brain\\Mets} & Mean \\
 & & \cite{bladder_cancer} & \cite{medseg_database} & \cite{qinbreast} & \cite{vallieres2015sarcoma} & \cite{kinahan2019acrin-nsclc} & \cite{isbi-ms} & \cite{adrenalacc} & \cite{hcctace} & \cite{stanfordBrainMets} & \\
Modality & & \MRI & \CT & \PET & \MRI & \PET & \MRI & \CT & \CT & \MRI & \\
\midrule
BiomedParse~\cite{biomedparse} & \xmark & 10.4 & 17.9 & 0.8 & 36.2 & 2.7 & 9.6 & 52.9 & 41.3 & 9.6 & 20.2 \\
Text3DSAM~\cite{xin2025text3dsam} & \cmark & 2.0 & 7.3 & 0.0 & 12.8 & 12.2 & 0.2 & 0.0 & 0.5 & 0.1 & 3.9 \\
SegVol~\cite{segvol} & \cmark & 0.1 & 32.2 & 0.1 & 0.8 & 0.0 & 0.0 & 51.1 & 58.4 & 0.0 & 15.9 \\
BiomedParse-V2~\cite{boltzformer} & \xmark & 2.7 & 16.6 & 0.0 & 7.7 & 0.5 & 2.0 & 55.9 & \textbf{70.4} & 18.7 & 19.4 \\
SAT~\cite{SAT} & \cmark & 9.6 & 0.0 & 58.3 & 10.6 & \textbf{77.1} & 13.7 & 0.1 & 62.3 & 22.2 & 28.2 \\
VoxTell~\cite{rokuss2025voxtell} & \cmark & 23.3 & \second{69.1} & \textbf{69.7} & 39.3 & \second{62.3} & \textbf{72.4} & \textbf{76.7} & \second{66.3} & \textbf{46.7} & \second{58.4} \\
\midrule
\oursrow \oursh & \xmark & 30.8 & 0.0 & 42.1 & 65.9 & 24.2 & 50.8 & 58.0 & 38.2 & 23.5 & 37.1 \\
\oursrow \ourss & \xmark & \second{41.8} & 0.0 & 54.2 & \second{74.4} & 34.7 & 54.7 & \second{65.9} & 51.7 & 23.5 & 44.5 \\
\oursrow \oursf & \xmark & \textbf{43.0} & \textbf{82.2} & \second{60.9} & \textbf{76.5} & 56.0 & \second{60.8} & 58.3 & 59.3 & \second{31.6} & \textbf{58.7} \\
 
\bottomrule
\end{tabular}
\end{adjustbox}
\caption{\small\textbf{Zero-shot 3D segmentation of tumors and lesions.} We show the Dice scores on several tumors and lesions benchmarks. The datasets by column are FedBCa (Center 2),
MedSeg Esophageal, QIN-Breast, Soft-Tissue-Sarcoma, ACRIN-NSCLC-FDG-PET, ISBI 2015 MS, Adrenal-ACC-Ki67-Seg, HCC-TACE-Seg, and 
BrainMetShare. Some baselines use 3D segmentation networks, while \ours uses a frozen, general-purpose MLLM (in parentheses), without any segmentation
network, or tools. The baselines other than VoxTell are quoted from~\cite{rokuss2025voxtell}, with VoxTell run by us on the benchmarks. Mean is over the
nine benchmarks. \textbf{Bold}: best, \underline{underline}: second best. \ours, with a strong backbone, matches and sometimes surpasses the trained state of the art on some modalities.}
\label{tab:main}
\end{table*}

\subsection{Planning the target from text}
Free text is turned into an explicit, inspectable plan before any localization. The MLLM $\mathbf{F}$ receives
$\mathbf{q}$, the modality, and three orthogonal overview slices of $\mathbf{V}$, and returns the plan. The plan specifies the
display window, the expected appearance, location, and size $s$ of the target, and likely
confounders. It also records its multiplicity (one structure or several parts,
such as metastases) and shape class (solid, shell, tubular, or branching). This plan selects the rendering, sets the sampling density of the later stages from $s$, and is quoted in
every subsequent prompt. The same model that reads the prompt thus also decides how to look for the target.

\subsection{Coarse-to-fine localization}
A volume holds hundreds of slices, and most of them do not show the target. The MLLM $\mathbf{F}$ therefore localizes the
target in two calls. In the \emph{coarse} call, it sees a sparse set of axial views, spaced according to $s$, plus one
coronal and one sagittal view for context. It returns a box in every view that shows the target, which gives an initial
extent along $z$. In the \emph{fine} call, it sees dense axial slices over this extent and returns one box
$\mathbf{B}_z$ per slice $z$. These boxes fix the extent $[z_{\min},z_{\max}]$ of the target and its in-plane region on
every slice. For a target with several parts, boxes are linked across slices into one track per part, each
traced separately.

\subsection{Keyframe contouring}
The MLLM traces the target only on keyframes $z_k$, a subset of the slices in $[z_{\min},z_{\max}]$ spaced according to
$s$, plus one probe slice beyond each end. Each keyframe is shown to $\mathbf{F}$ as a \emph{pair} of images. The full
slice, with the crop outlined, lets it recognize the target in context. The zoomed crop around $\mathbf{B}_{z_k}$ is where
it traces the boundary. The MLLM either answers ``absent'' or returns closed loops of points, one per connected part, with optional
hole loops. The curve through the $n$ points $\mathbf{p}_j=(u_j,v_j)$ of a loop is the closed interpolating cubic spline \cite{hamdi2026llmseg}
\begin{equation}
  \mathbf{C}_k(t)=\sum_{j=0}^{n-1} \mathbf{c}_j\,N_3\!\left(t-j\right),\quad \mathbf{C}_k(j)=\mathbf{p}_j,
  \label{eq:spline}
\end{equation}
with $t\in[0,n)$, periodic cubic B-spline basis $N_3$, and control points $\mathbf{c}_j$ solved so that the curve passes through
the points. The curves $\mathbf{C}_k$ enclose the slice region $\Omega_k$, defined as the interior of the loops minus any
holes. The keyframe mask is its indicator, $\mathbf{m}_k(\mathbf{u})=[\mathbf{u}\in\Omega_k]$, where $[\cdot]$ is $1$ if
the statement holds and $0$ otherwise. When $\mathbf{F}$ marks a probe slice as ``present'', the extent grows by another
probe slice. Each call holds only a few keyframes, because the MLLM draws coarser outlines and drops parts as the number
of images per call grows.

\subsection{3D assembly}
Each keyframe mask becomes a signed distance map $\boldsymbol{\phi}_k$. It holds the distance in mm from every pixel
$\mathbf{u}$ of the slice to the traced boundary $\partial\Omega_k$, negative inside,
\begin{equation}
  \boldsymbol{\phi}_k(\mathbf{u})=\bigl(1-2\,\mathbf{m}_k(\mathbf{u})\bigr)\min_{\mathbf{u}'\in\partial\Omega_k}\lVert\mathbf{u}-\mathbf{u}'\rVert_2 .
  \label{eq:dist}
\end{equation}
Between consecutive keyframes, the volume is filled by linear blending of these maps,
\begin{equation}
  \boldsymbol{\phi}(\mathbf{u},z)=(1-\lambda)\,\boldsymbol{\phi}_k(\mathbf{u})+\lambda\,\boldsymbol{\phi}_{k+1}(\mathbf{u}),
  \label{eq:sdf}
\end{equation}
with $\lambda=(z-z_k)/(z_{k+1}-z_k)$. The assembled mask is the negative region of the blended field,
$\widetilde{\mathbf{M}}(\mathbf{u},z)=[\boldsymbol{\phi}(\mathbf{u},z)<0]$. The blend morphs the shape smoothly from one
keyframe to the next. Where a present keyframe neighbors an absent one, the mask ends halfway between them.

\mysection{Intensity snapping}
A final training-free step snaps the boundary of $\widetilde{\mathbf{M}}$ to the image $\mathbf{V}$. Medical images are acquired under energy bands that increase contrast between normal and abnormal tissue.
Let $d(\mathbf{u},z)$ be the distance in mm of a voxel to this boundary and $b$ the half-width of a band around it. The
mask's interior beyond the band has median intensity $\mu_{\mathrm{in}}$ and spread $\sigma_{\mathrm{in}}$. A shell
outside the band has median $\mu_{\mathrm{out}}$ and spread $\sigma_{\mathrm{out}}$. A band voxel takes the side
$c\in\{\mathrm{in},\mathrm{out}\}$ of lower cost
$\ell_c=\lvert\mathbf{V}(\mathbf{u},z)-\mu_c\rvert/\sigma_c+\log\sigma_c$,
\begin{equation}
  \mathbf{M}(\mathbf{u},z)=
  \begin{cases}
    \bigl[\ell_{\mathrm{in}}<\ell_{\mathrm{out}}\bigr], & d(\mathbf{u},z)\le b,\\[2pt]
    \widetilde{\mathbf{M}}(\mathbf{u},z), & \text{otherwise.}
  \end{cases}
  \label{eq:snap}
\end{equation}

\begin{figure*}[t]
  \centering
  \setlength{\qgw}{0.116\textwidth}
  \setlength{\tabcolsep}{0.5pt}
  \renewcommand{\arraystretch}{0.4}
  \begin{tabular}{@{}c*{3}{c}@{\hspace{3pt}}*{3}{c}@{\hspace{3pt}}*{2}{c}@{}}
   & \multicolumn{3}{c}{\qhd{Axial slice 1}} & \multicolumn{3}{c}{\qhd{Axial slice 2}} & \multicolumn{2}{c}{\qhd{3D}} \\[2pt]
   & \qhd{Image} & \qhd{Ground truth} & \qhd{\ours (ours)} & \qhd{Image} & \qhd{Ground truth} & \qhd{\ours (ours)} & \qhd{Ground truth} & \qhd{\ours (ours)} \\
  \qrow{AMOS (MRI)}{15 organs} & \qimh{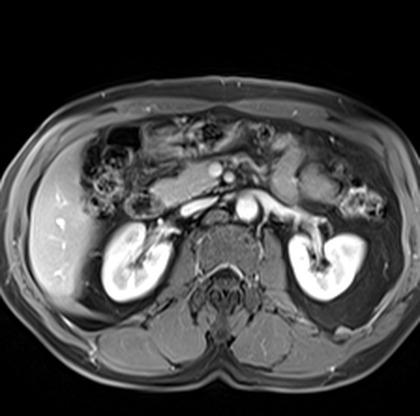} & \qimh{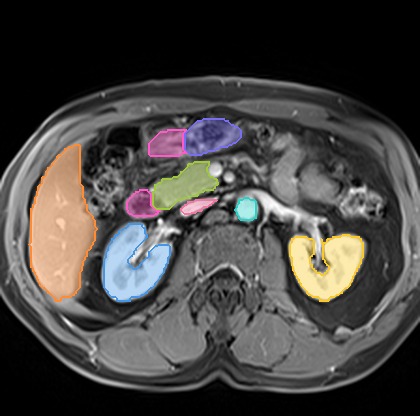} & \qimh{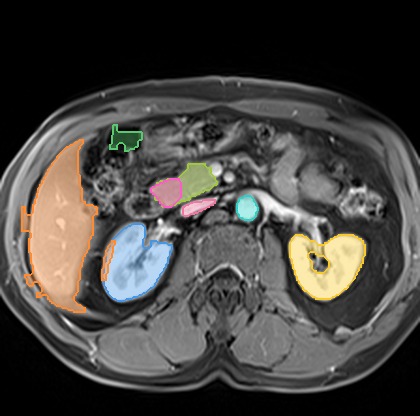} & \qimh{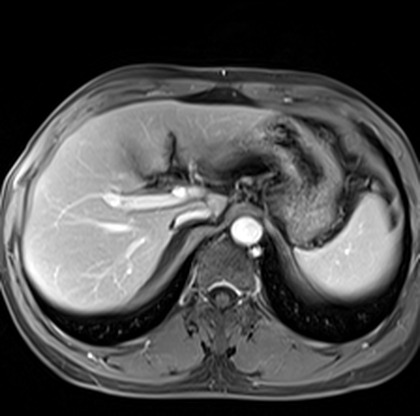} & \qimh{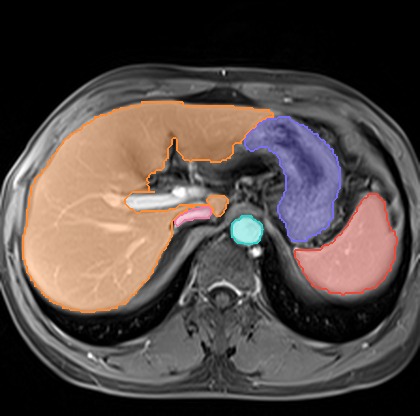} & \qimh{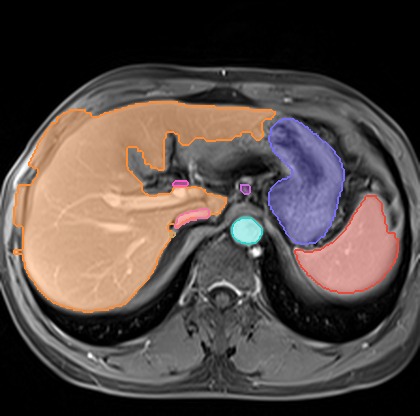} & \qimh{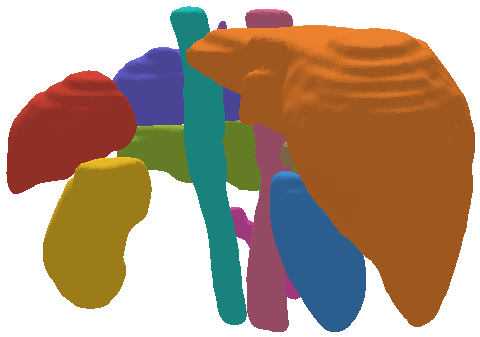} & \qimh{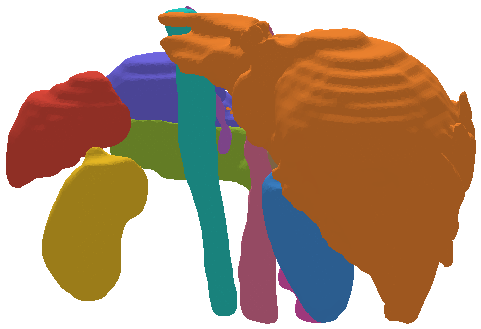} \\
  \qrow{HVSMR (MRI)}{8 heart structures} & \qimh{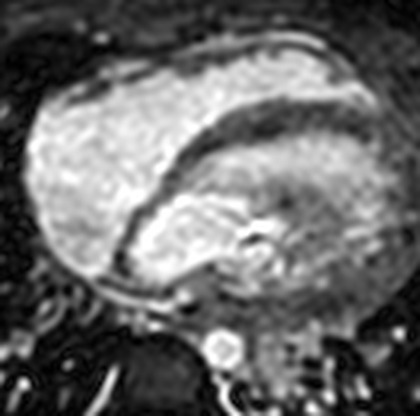} & \qimh{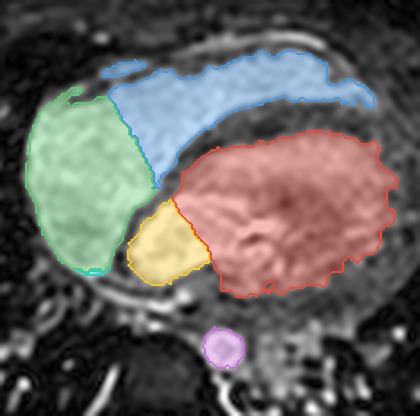} & \qimh{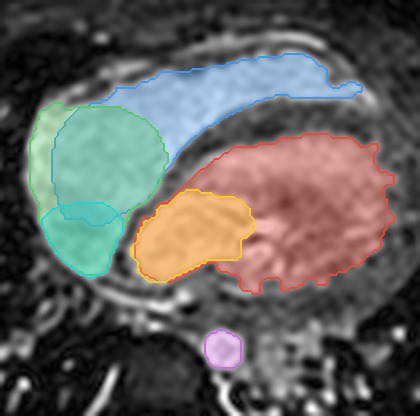} & \qimh{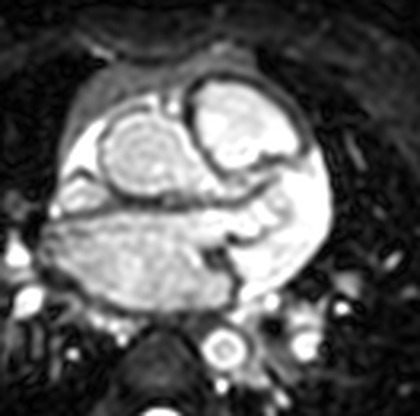} & \qimh{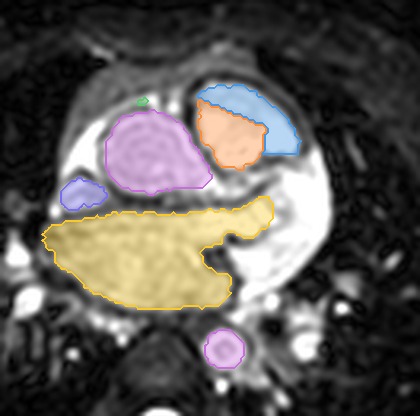} & \qimh{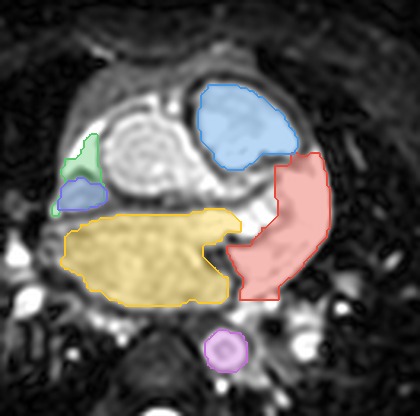} & \qimh{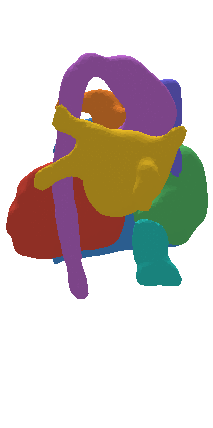} & \qimh{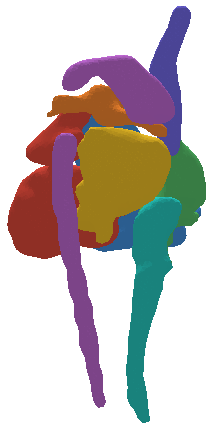} \\
  \qrow{PENGWIN (CT)}{sacrum, hip bones} & \qimh{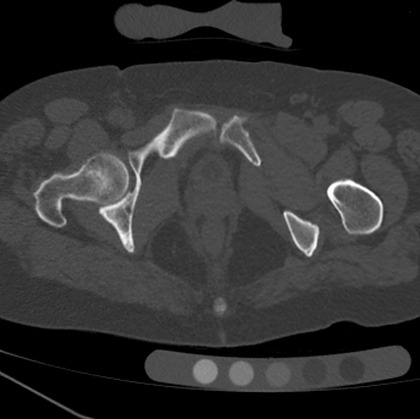} & \qimh{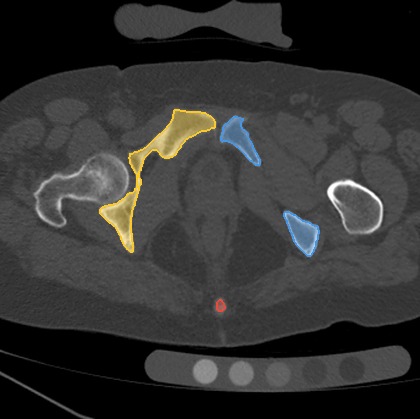} & \qimh{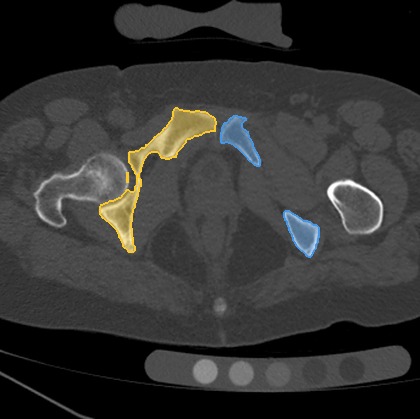} & \qimh{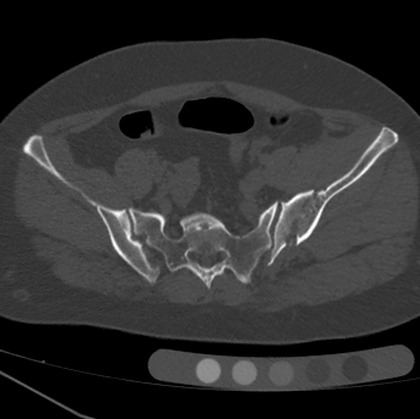} & \qimh{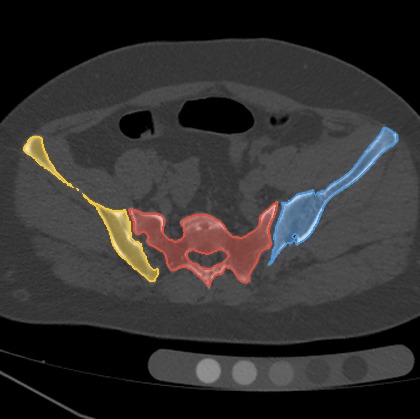} & \qimh{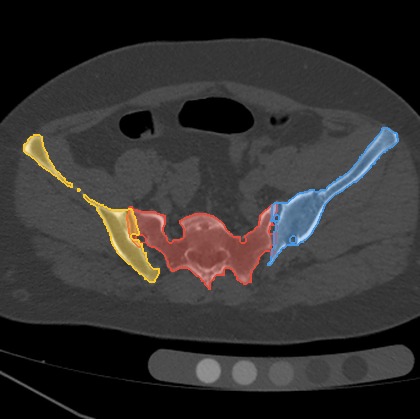} & \qimh{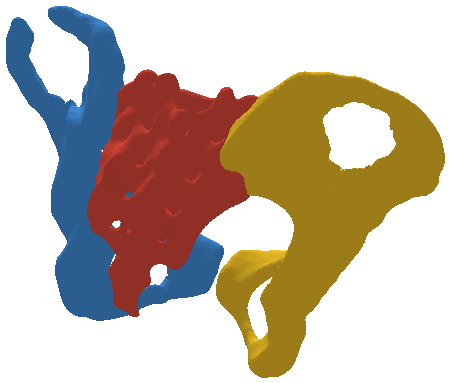} & \qimh{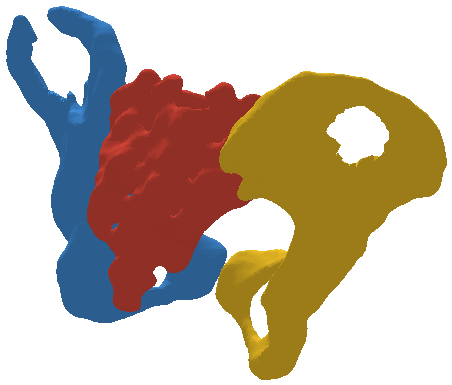} \\
  \qrow{AeroPath (CT)}{lung, trachea} & \qimh{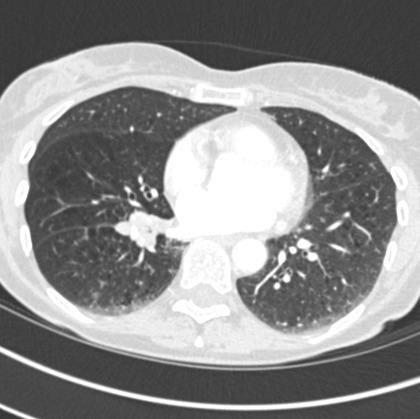} & \qimh{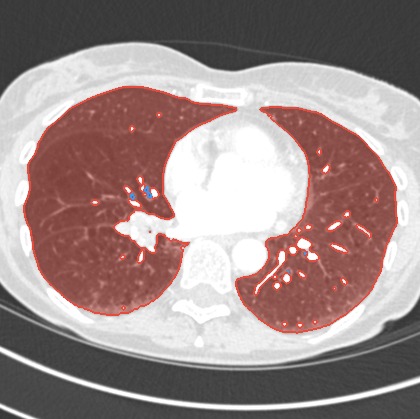} & \qimh{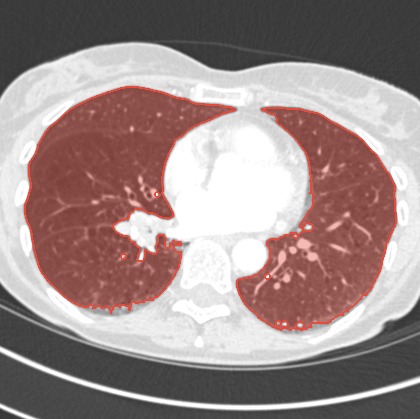} & \qimh{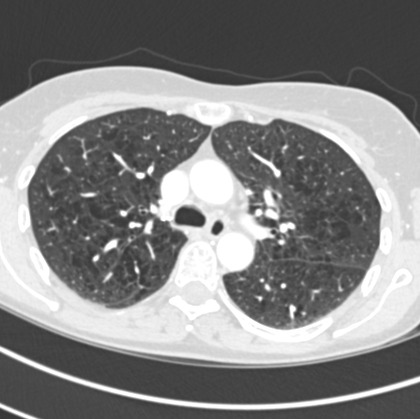} & \qimh{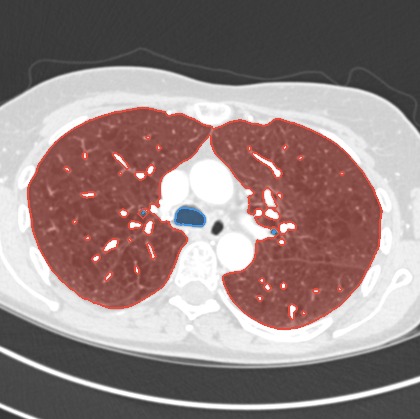} & \qimh{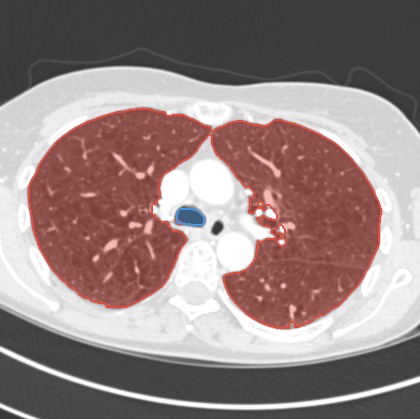} & \qimh{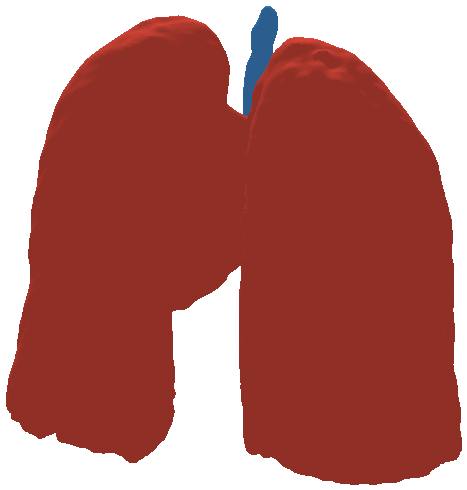} & \qimh{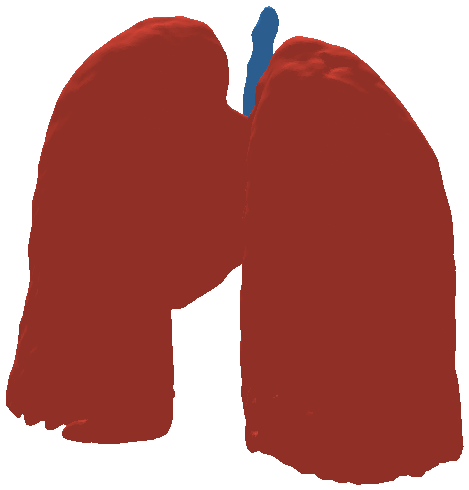} \\
  \qrow{VEELA (CT)}{liver, vessels} & \qimh{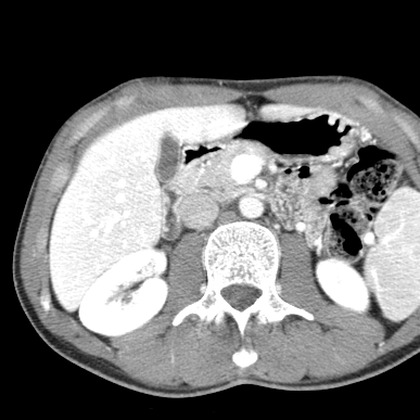} & \qimh{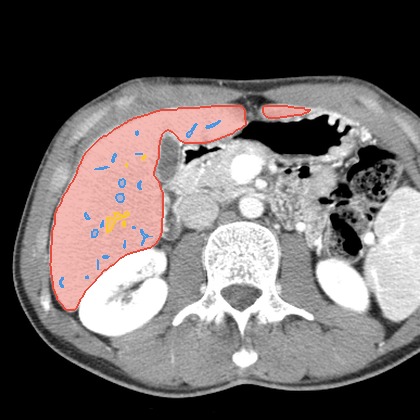} & \qimh{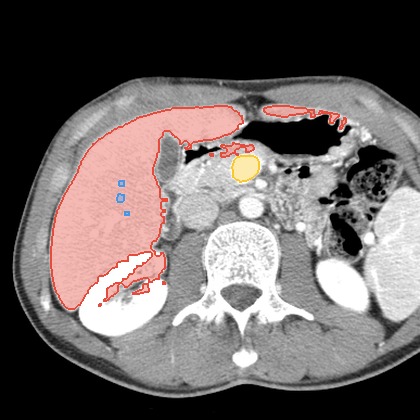} & \qimh{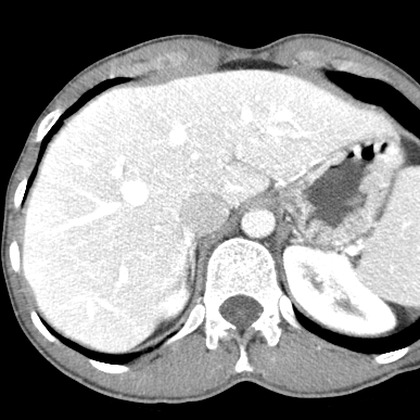} & \qimh{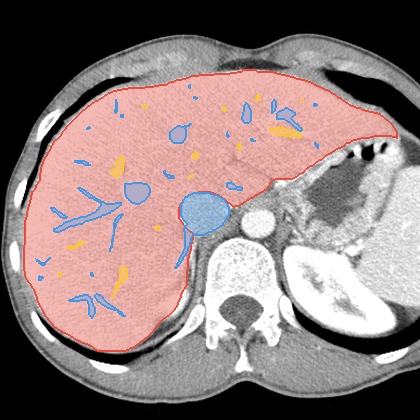} & \qimh{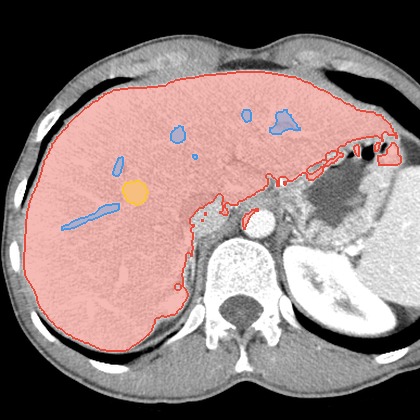} & \qimh{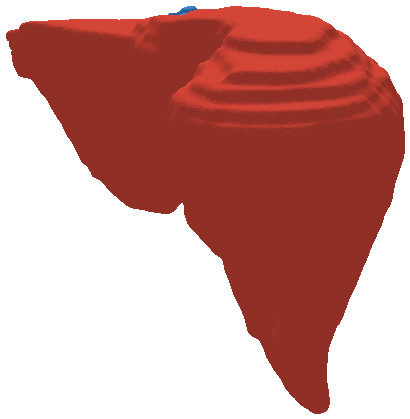} & \qimh{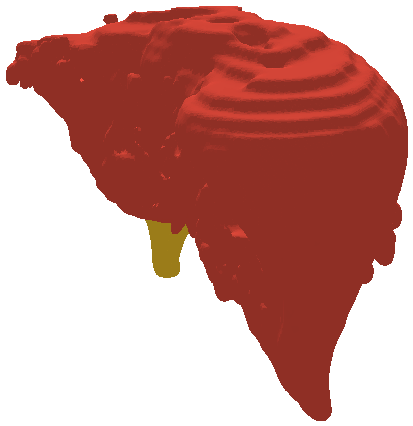} \\
  \end{tabular}
  \caption{\small\textbf{Qualitative results on anatomical structures.} Each row shows one representative
  case of a benchmark on two axial slices and as 3D surfaces. Each target keeps one color across all columns.}
  \label{fig:qual_anat}
\end{figure*}

\section{Experiments} \label{sec:experiments}

\mysection{Benchmarks}
We evaluate on 17 public benchmarks (923 volumes, 79 prompts) and follow the protocol of VoxTell~\cite{rokuss2025voxtell} on
all its publicly available held-out benchmarks. Nine hold tumors and lesions. They test unknown
concepts~\cite{bladder_cancer,medseg_database}, known concepts in a new modality~\cite{qinbreast,vallieres2015sarcoma},
and unseen datasets of known
concepts~\cite{kinahan2019acrin-nsclc,isbi-ms,adrenalacc,hcctace,stanfordBrainMets}. Five hold
anatomy~\cite{pediatric-ct-seg,2023aeropath,toprak2025veela,pace2024hvsmr,pengwin}. Three are standard
3D \emph{semantic segmentation} benchmarks~\cite{BTCV,AMOS,Baid2021TheRB}, scored as in~\cite{Bourigault_2025_ICCV} on ten shared
abdominal organs and 100 seeded BraTS cases.

\mysection{Protocol and baselines}
Every method receives the benchmark's label name as its prompt, with no tuning. We report Dice $\times 100$, averaged
over cases per prompt, then over prompts. Zero-shot means \ours uses no label, example, or tuning from any
benchmark. The training data of its MLLM backbones is not public. We quote published numbers of the trained
text-promptable models~\cite{biomedparse,xin2025text3dsam,segvol,boltzformer,SAT} and the trained 3D
segmenters~\cite{ronneberger2015unet,VNet,ResUNet++,AttentionUNet,nnFormer,UNETR,SegResNet,Hatamizadeh2022SwinUS,graf2024totalvibesegmentator,Bourigault_2025_ICCV}.
For VoxTell~\cite{rokuss2025voxtell}, we run the released checkpoint on our exact cases. Benchmark details and complete tables are in the \supp.

\mysection{Backbones and implementation}
We use three backbones from one family with identical prompts and settings, namely \haiku, \sonnet~\cite{anthropic2026claude55},
and \fable, in order of increasing strength. The two smaller ones run on every benchmark. \fable costs about ten times
more per volume than \sonnet, so it runs on the tumor and lesion benchmarks and BraTS only. In
\secLabel\ref{sec:results}, \ours without a backbone name refers to the strongest backbone in that table. The qualitative
figures show \fable on tumors and lesions and \sonnet on anatomy.
One set of settings serves every benchmark, modality, and backbone. Coarse localization shows 16 to 48 axial views at
most $s/2$ apart. Fine localization shows slices 8~mm apart, at most 24 per call. Keyframes are spaced
$\min(8\,\mathrm{mm},s/2)/3$ apart, with at most 8 per contour call, and the extent grows by at most three
probes. Views are rendered at 768 pixels and crops at 1024 pixels on the long side. Each loop has 8 to 40 points. The
snapping band $b$ is 0.08 times the mask's median cross-sectional diameter, clipped between 1 and 8~mm. Snapping is
skipped when the inside and outside medians lie less than 0.5 pooled spreads apart. Plan and localization calls use
medium reasoning effort, and contour calls use the highest available effort. All settings were chosen on development
cases of the anatomical benchmarks and confirmed on held-out ones (\secLabel\ref{sec:analysis}).

\section{Results} \label{sec:results}

\mysection{Tumors and lesions}
Without any additional training, \ours reaches 58.7 mean Dice over the nine benchmarks of \tblLabel\ref{tab:main}. This is level
with the best trained model on identical cases (58.4) and far above all other baselines (at most 28.2). \ours leads on
unknown concepts and new modalities (65.7 vs.\ 50.3), most clearly on sarcoma (76.5 vs.\ 39.3). The trained model leads
on unseen datasets of known concepts (64.9 vs.\ 53.2), especially on small brain metastases. With no change to the pipeline,
the mean rises from 37.1 to 44.5 and 58.7 across the three backbones.

\begin{table}[t]
\centering
\setlength{\tabcolsep}{2.6pt}
\begin{adjustbox}{max width=\columnwidth}
\begin{tabular}{lc|ccccc|c}
\toprule
Method & \shortstack{3D Seg.\\Network} & \shortstack{Organs\\(pediatric)} & \shortstack{Lung \&\\Airway} & \shortstack{Liver \&\\Vessels} & Heart & \shortstack{Pelvic\\Bones} & Mean \\
 & & \cite{pediatric-ct-seg} & \cite{2023aeropath} & \cite{toprak2025veela} & \cite{pace2024hvsmr} & \cite{pengwin} & \\
Modality & & \CT & \CT & \CT & \MRI & \CT & \\
\midrule
TotalSegmentator~\cite{wasserthal2023totalsegmentator} & \cmark & 67.6 & 85.2 & \second{49.5} & \na & \na & \na \\
BiomedParse~\cite{biomedparse} & \xmark & 9.1 & 0.0 & 6.5 & 8.1 & 0.5 & 4.8 \\
Text3DSAM~\cite{xin2025text3dsam} & \cmark & 26.6 & 65.4 & 28.9 & 15.7 & 24.8 & 32.3 \\
SegVol~\cite{segvol} & \cmark & 52.5 & \second{88.7} & 41.3 & 1.9 & 22.1 & 41.3 \\
BiomedParse-V2~\cite{boltzformer} & \xmark & 51.8 & 62.6 & 35.1 & 18.1 & 1.4 & 33.8 \\
SAT~\cite{SAT} & \cmark & \second{68.8} & 88.0 & 37.7 & 43.4 & \second{96.1} & \second{66.8} \\
VoxTell~\cite{rokuss2025voxtell} & \cmark & \textbf{72.8} & \textbf{89.7} & \textbf{57.2} & \textbf{52.7} & \textbf{97.6} & \textbf{74.0} \\
\midrule
\oursrow \oursh & \xmark & 56.2 & 84.6 & 44.7 & 40.4 & 87.0 & 62.6 \\
\oursrow \ourss & \xmark & 62.3 & 84.1 & 47.5 & \second{49.0} & 89.8 & 66.5 \\
 
\bottomrule
\end{tabular}
\end{adjustbox}
\caption{\small\textbf{Zero-shot 3D segmentation of anatomical structures.} We show the Dice scores on five anatomy
benchmarks with multiple organs and parts. The datasets by column are Pediatric-CT-SEG, AeroPath, VEELA, HVSMR-2.0, and PENGWIN. Baselines and symbols are as in
\tblLabel\ref{tab:main}. \ours (gray rows) is run here with its two smaller backbones. Mean is over the five benchmarks. \ours does not use 3D segmentation networks or tools, but rivals those baselines.}
\label{tab:anatomy}
\end{table}

\mysection{Anatomical structures}
With \sonnet, \ours reaches 66.5 mean Dice on anatomy (\tblLabel\ref{tab:anatomy}). This is level with the second-best
trained model (66.8) and 7.5 points below the best. On large structures such as the lungs, the liver, and the pelvic bones,
\ours is close to the trained models. The gap lies in small and thin structures. Running \fable on these anatomy
benchmarks, as well as on BTCV and AMOS, is cost-prohibitive, since each volume holds numerous organs.

\mysection{3D semantic segmentation and the value of training}
Every baseline in \tblLabel\ref{tab:standard} has seen voxel labels of exactly these structures. \ours reaches 76.7
and 69.9 on the abdominal organs of BTCV~\cite{BTCV} and AMOS~\cite{AMOS}, and 73.9 on the brain tumors of
BraTS~\cite{Baid2021TheRB}, from the class names alone. In-domain
training is thus worth about 12 to 17 points over a frozen MLLM here.

\begin{table}[t]
\centering
\setlength{\tabcolsep}{5pt}
\renewcommand{\arraystretch}{0.92}
\begin{adjustbox}{max width=\columnwidth}
\begin{tabular}{lc|cc|c}
\toprule
Method & \shortstack{3D Seg.\\Network} & BTCV~\cite{BTCV} & AMOS~\cite{AMOS} & BraTS~\cite{Baid2021TheRB} \\
Modality & & \CT & \MRI & \MRI \\
\midrule
UNet~\cite{ronneberger2015unet} & \cmark & \na & \na & 54.4 \\
V-Net~\cite{VNet} & \cmark & \na & \na & 84.2 \\
ResUNet++~\cite{ResUNet++} & \cmark & \na & \na & 78.4 \\
Attention U-Net~\cite{AttentionUNet} & \cmark & \na & \na & 79.8 \\
nnFormer~\cite{nnFormer} & \cmark & \na & \na & 81.2 \\
UNETR~\cite{UNETR} & \cmark & \na & \na & 87.1 \\
SegResNet~\cite{SegResNet} & \cmark & \na & \na & 89.0 \\
Swin-UNETR~\cite{Hatamizadeh2022SwinUS} & \cmark & \na & \na & 88.6 \\
TotalVibeSeg.~\cite{graf2024totalvibesegmentator} & \cmark & 79.2 & 79.6 & 83.0 \\
Swin-BOB~\cite{Bourigault_2025_ICCV} & \cmark & 88.2 & 84.0 & 89.4 \\
VoxTell~\cite{rokuss2025voxtell} & \cmark & 88.2 & 86.4 & 88.4 \\
\midrule
\oursrow \oursh & \xmark & 71.8 & 62.7 & 62.2 \\
\oursrow \ourss & \xmark & 76.7 & 69.9 & 73.9 \\
 
\bottomrule
\end{tabular}
\end{adjustbox}
\caption{\small\textbf{In-Domain 3D semantic segmentation.} We show the mean Dice scores on three standard benchmarks. BTCV~\cite{BTCV} and AMOS~\cite{AMOS}
are multi-organ benchmarks, scored on their
ten shared abdominal organs. BraTS~\cite{Baid2021TheRB} is a multi-region brain tumor benchmark, scored on its three tumor regions. Every
baseline was trained on voxel labels of the same structures. All BraTS baselines are supervised on
BraTS~\cite{Bourigault_2025_ICCV} (VoxTell on a later release). On BTCV and AMOS, TotalVibeSeg.\ and Swin-BOB are
trained on large labeled collections of the same organs, while VoxTell is trained on BTCV and AMOS. \ours
prompts each class by its name and uses no 3D labels or tools.}
\label{tab:standard}
\end{table}

\begin{figure*}[t]
  \centering
  \setlength{\qgw}{0.138\textwidth}
  \setlength{\tabcolsep}{0.6pt}
  \renewcommand{\arraystretch}{0.4}
  \begin{tabular}{@{}*{4}{c}@{\hspace{6pt}}*{3}{c}@{}}
   \multicolumn{4}{c}{\qhd{2D}} & \multicolumn{3}{c}{\qhd{3D}} \\[1pt]
   \cmidrule(lr){1-4}\cmidrule(lr){5-7}
   \qhd{Image} & \qhd{Ground truth} & \qhd{\ours (ours)} & \qhd{VoxTell~\cite{rokuss2025voxtell}} & \qhd{Ground truth} & \qhd{\ours (ours)} & \qhd{VoxTell~\cite{rokuss2025voxtell}} \\
  \qimh{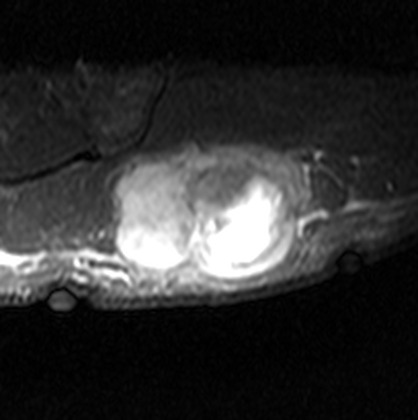} & \qimh{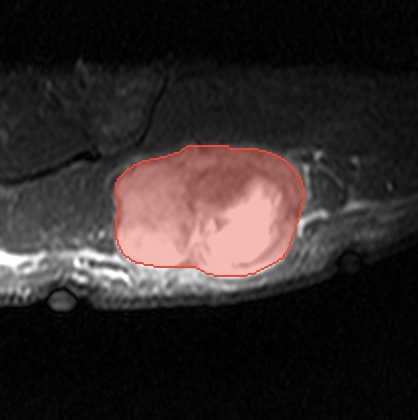} & \qimh{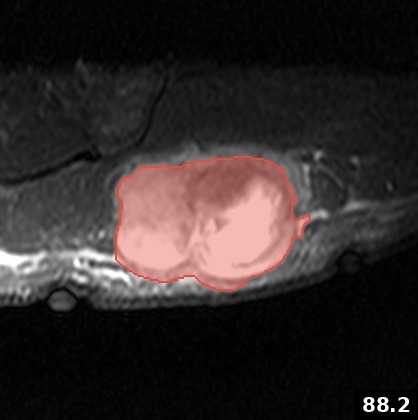} & \qimh{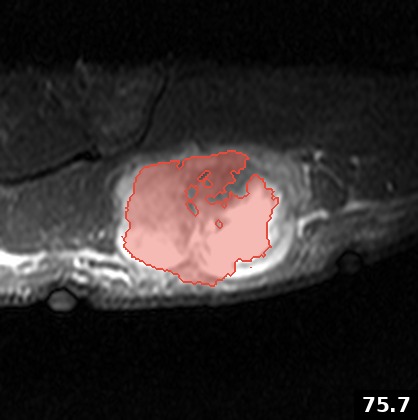} & \qimh{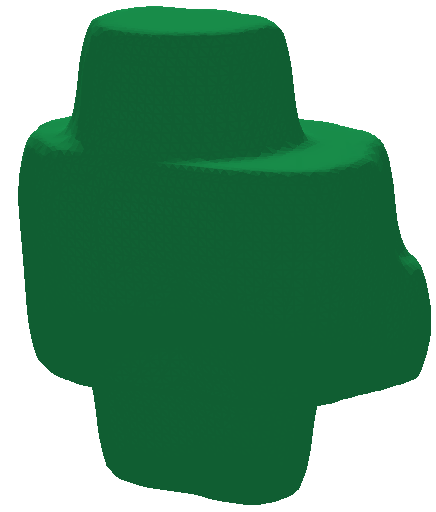} & \qimh{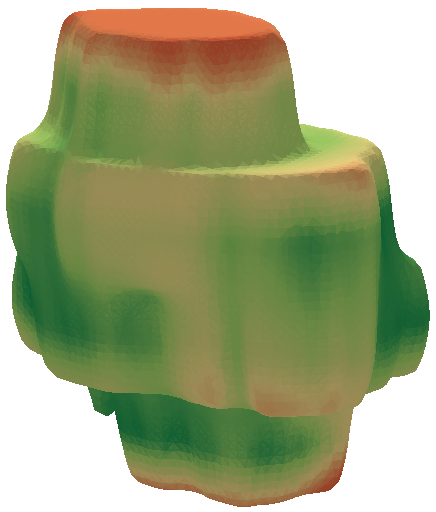} & \qimh{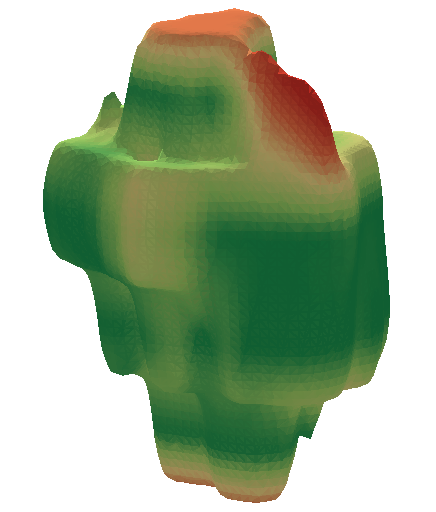} \\
  \qimh{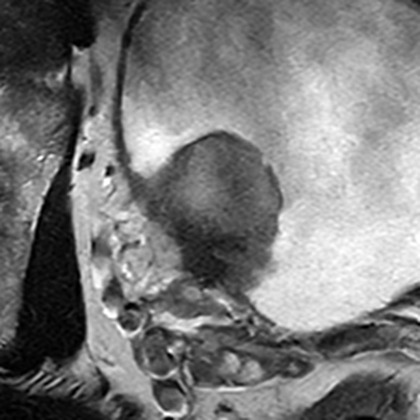} & \qimh{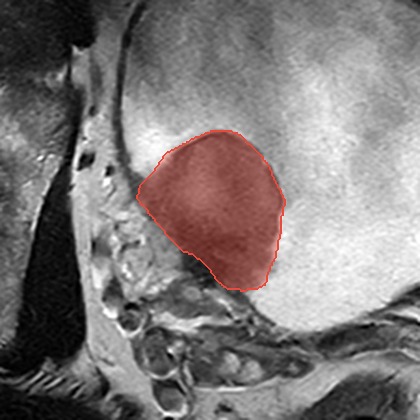} & \qimh{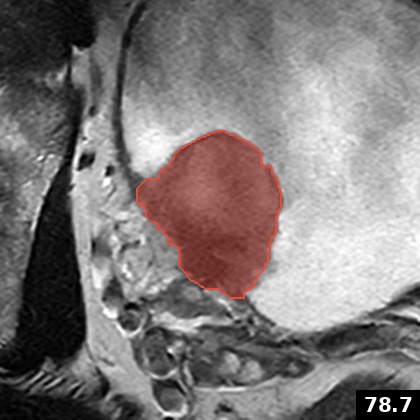} & \qimh{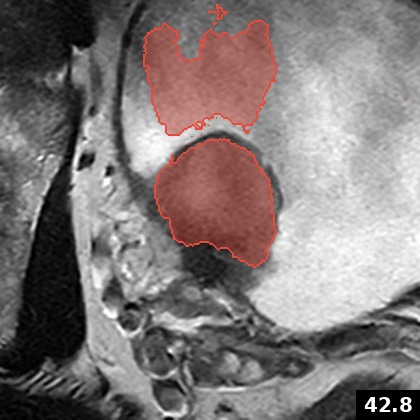} & \qimh{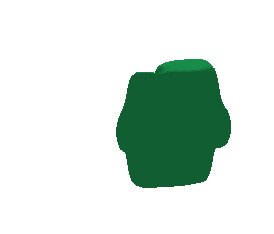} & \qimh{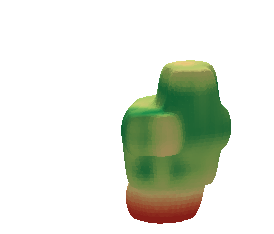} & \qimh{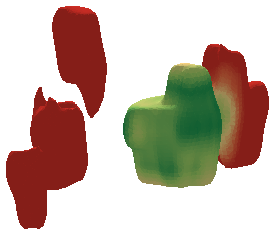} \\
  \qimh{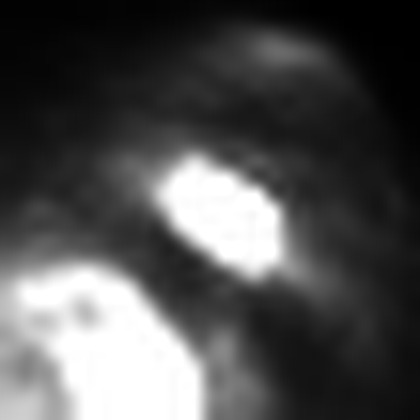} & \qimh{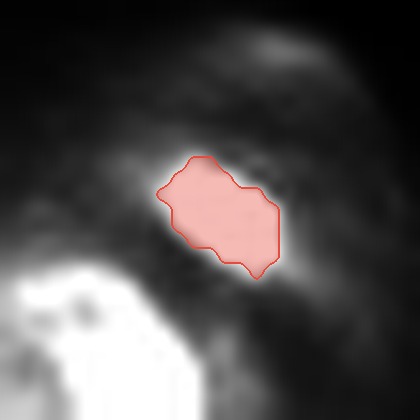} & \qimh{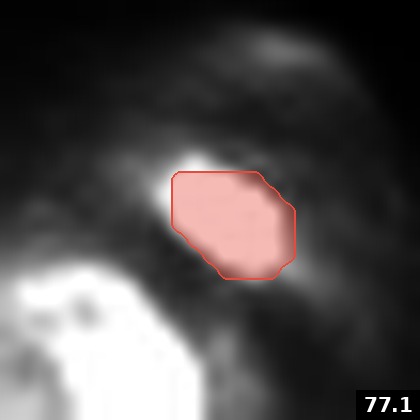} & \qimh{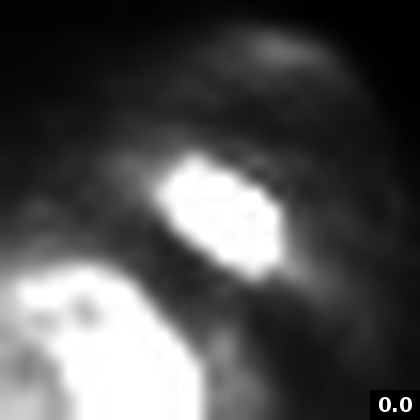} & \qimh{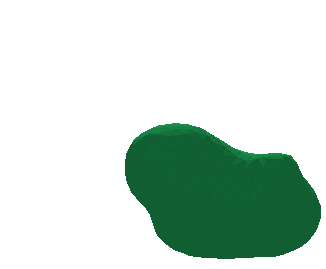} & \qimh{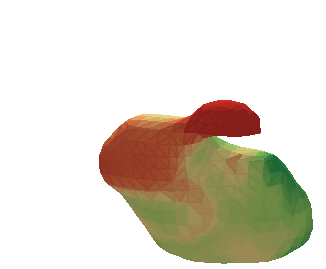} & \qimh{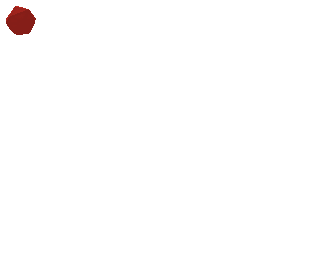} \\
   & & & & \multicolumn{3}{c}{\includegraphics[width=2.6\qgw]{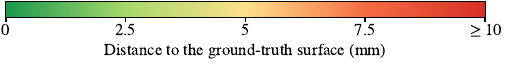}} \\
  \end{tabular}
  \caption{\small\textbf{Qualitative comparisons on tumors and lesions.} The rows show, from top to bottom, a sarcoma on T2
  MRI (prompt ``soft tissue lesion''), a bladder carcinoma on T2 MRI (FedBCa, ``carcinoma''), and a breast cancer on PET
  (QIN, ``fdg-avid breast cancer''). Each row is one representative case, shown on a 2D slice and as 3D surfaces next
  to the best trained model~\cite{rokuss2025voxtell}. In the 2D slices, the corner box numbers give the Dice of the shown targets.
  In 3D, each predicted surface is colored by its distance to the ground-truth surface, from green (on it) to red
  (10~mm or more away). On sarcoma and bladder carcinoma, which trained models did not see in this form, the frozen MLLM recognizes the lesion from the
  text alone.}
  \label{fig:qual_path}
\end{figure*}

\mysection{Qualitative results}
\figLabel\ref{fig:qual_anat} and \figLabel\ref{fig:qual_path} show representative cases, and the \supp adds best and
failure cases. Typical errors are missed thin ends, leaks into neighbors of similar intensity, and missed small
lesions.

\section{Analysis} \label{sec:analysis}

\begin{figure}[t]
  \centering
  \includegraphics[width=\linewidth]{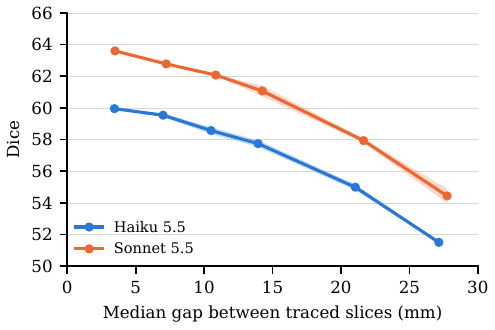}
  \vspace{-14pt}
  \caption{\small \textbf{How many slices must the MLLM trace?} Dice on held-out anatomy volumes against the median gap
  between traced slices when only a subset of each target's keyframes is kept and its mask is reassembled.}
  \label{fig:density}
  \vspace{-6pt}
\end{figure}

\tblLabel\ref{tab:design} removes or replaces one design choice at a time, and
\figLabel\ref{fig:density} varies the density of the traced keyframes. All rows use \haiku on held-out volumes.
Setups and more analyses are in the \supp.

\mysection{Design choices}
On anatomy, every stage contributes. The plan and fine localization matter most. Removing either costs about three
to four Dice points, mainly on heart structures that lie close together. Tracing benefits from larger crops and
higher reasoning effort, and intensity snapping adds 2.4 points without any MLLM call. Distance blending smooths the
surface between keyframes without changing the overlap.

\begin{figure*}[t]
  \centering
  \includegraphics[width=\linewidth]{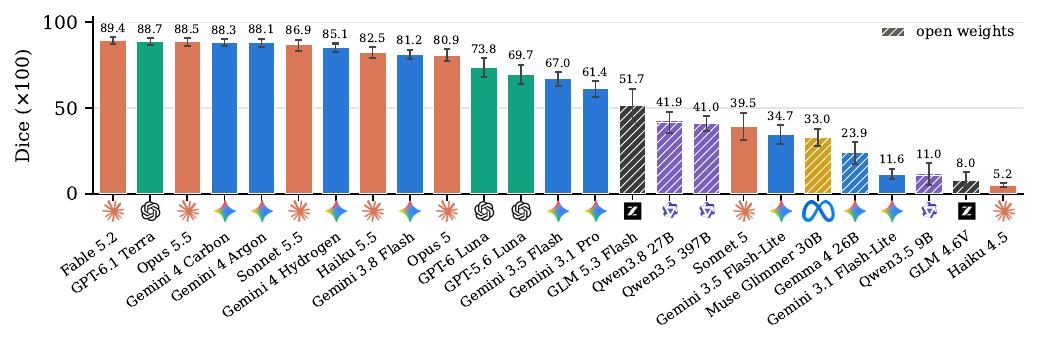}
  \vspace{-6pt}
  \caption{\small\textbf{The quality of \ours follows the quality of its backbone.} Dice of the unchanged pipeline
  with 25 backbone MLLMs on the same ten held-out PENGWIN volumes. Whiskers are 95\% bootstrap intervals over volumes. Open-weights models are drawn with thin white diagonal
  stripes.}
  \label{fig:backends}
  \vspace{-4pt}
\end{figure*}

\mysection{The backbone sets the accuracy}
\ours improves with the backbone without any change to the pipeline. On the full benchmarks, \fable is better than
\sonnet on 8 of the 9 tumor and lesion benchmarks, by 14.2 Dice on average. Beyond one model family, we run the
unchanged pipeline with 25 backbones from six providers on the ten pelvic CT volumes of \figLabel\ref{fig:slice}
(\figLabel\ref{fig:backends}). Dice spans 5 to 89 and follows the model generation, not the provider. The newest
models of three providers are within 1.3 points of each other, and open-weights models still score far below them.
Within every model line, the newer generation is more accurate, so \ours improves with no work on our side.

\mysection{Keyframe density}
The MLLM traces only a subset of the slices, and geometry fills the rest. \figLabel\ref{fig:density} shows how sparse
this subset can be. For both backbones, Dice changes by less than one point up to a gap of 7~mm between traced slices
and falls quickly beyond 14~mm. The default gap of 3.4~mm (the leftmost points) thus lies on the flat part of the
curve. Tracing three times fewer keyframes would save over a third of the calls at the cost of one Dice point. Denser
keyframes (1.7 times as many as the default) changed Dice by only +0.3 (s.e.\ 0.4) with \haiku, at extra MLLM calls.

\mysection{Trained tracer}
Replacing the MLLM tracer with MedSAM2~\cite{MedSAM2}, prompted by the boxes of our fine localization, cuts the calls
about sevenfold but loses 3.3 points.

\mysection{Running MLLMs for each slice}
The direct way to segment a volume with an MLLM is to run a 2D MLLM segmenter~\cite{hamdi2026llmseg} on every slice
and stack the masks. On the ten pelvic CT volumes of \figLabel\ref{fig:slice}, this reaches 54.8 Dice with 4{,}182
calls per volume, while \ours reaches 82.5 Dice with 34 calls and the same MLLM. The gain of \ours thus comes from
planning, 3D localization, and assembly, not from a better 2D tracer.

\mysection{Recognition is the bottleneck}
The MLLM accurately traces what it recognizes. Of the Dice lost by the default pipeline on held-out anatomy volumes, only 20\%
comes from imperfect outlines inside a correct box. The rest comes from slices where the mask is off the target or
outside its extent. Likewise, a trained segmenter reaches 82.0 Dice with ground-truth boxes but only 48.7 with the
MLLM's boxes. The remaining errors are thus recognition errors on small, atypical, or low-contrast structures.

\begin{table}[t]
  \centering
  \small
  \setlength{\tabcolsep}{4pt}
  \begin{adjustbox}{width=0.9\linewidth}
  \begin{tabular}{@{}l cc@{}}
  \toprule
  \textbf{Configuration} & Dice on anatomy $\uparrow$ & Calls $\downarrow$ \\
  \midrule
  w/o plan (fixed defaults) & 57.2\,\sd{0.9} & 71 \\
\midrule
w/o fine localization & 56.1\,\sd{1.1} & 56 \\
\midrule
3$\times$ sparser keyframes & 59.0\,\sd{0.3} & 43 \\
512-pixel crops & 58.5\,\sd{0.5} & 72 \\
Lower reasoning effort & 58.2\,\sd{0.7} & 65 \\
\midrule
w/o distance blending & 59.9\,\sd{0.2} & 69 \\
w/o intensity snapping & 57.6\,\sd{0.5} & 69 \\
\midrule
MedSAM2~\cite{MedSAM2} as tracer & 56.7\,\sd{1.1} & 10 \\
\midrule
\oursrow \ours (default) & \textbf{60.0} & 69 \\
 
  \bottomrule
  \end{tabular}
  \end{adjustbox}
  \caption{\small \textbf{Ablation study of \ours.} Mean Dice and mean MLLM calls per volume with \haiku on anatomy when one
  component of \ours is removed or replaced. Rows are grouped by pipeline stage (plan, localization, contouring, and
  assembly) and an alternative tracer. All rows are paired on 36 held-out volumes from six anatomy benchmarks. $\pm$ is
  the standard error of the paired change against the default. Gray marks the default, and bold marks the best value.}
  \label{tab:design}
  \vspace{-6pt}
\end{table}

\mysection{Free-text prompts}
\ours reads the prompt as language rather than as a class label (\tblLabel\ref{tab:wording_main}). On anatomy, it
keeps its accuracy under synonyms and plain-language descriptions, while the best trained model falls from 91.6 to
65.9 Dice with descriptions. On tumors, a more explicit wording helps \ours most. Asking for ``bladder cancer'' instead
of ``carcinoma'' lifts it from 28.2 to 66.8, while the best trained model stays below 30 with every wording. The MLLM thus uses the
context of a prompt, not only its label, which helps where a concept is ambiguous or uncommon.

\begin{table}[t]
  \centering
  \small
  \setlength{\tabcolsep}{3pt}
  \begin{adjustbox}{max width=\linewidth}
  \begin{tabular}{@{}l ccc ccc@{}}
  \toprule
  & \multicolumn{3}{c}{Anatomy} & \multicolumn{3}{c}{Tumors} \\
  \cmidrule(lr){2-4}\cmidrule(l){5-7}
  \textbf{Method} & Label & Synonym & Descr. & Label & Synonym & Descr. \\
  \midrule
VoxTell~\cite{rokuss2025voxtell} & \textbf{91.6} & \textbf{91.3} & 65.9 & 29.7 & 28.0 & 20.6 \\
\oursrow \ours & 78.3 & 78.0 & \textbf{78.2} & \textbf{48.0} & \textbf{72.8} & \textbf{68.3} \\
 
  \bottomrule
  \end{tabular}
  \end{adjustbox}
  \vspace{-4pt}
  \caption{\small \textbf{Prompt wording.} Mean Dice when the same targets are named by the dataset label, a synonym,
  or a plain-language description. Anatomy covers bones, and abdominal organs across three benchmarks.
  Tumors cover soft-tissue sarcoma and bladder carcinoma. \ours is more robust to the text prompt variations than the best trained model.}
  \label{tab:wording_main}
  \vspace{-6pt}
\end{table}

\mysection{Cost}
Summed over all MLLM calls, a volume costs about \$0.14 with \haiku and \$1.5 with \sonnet on average, and \fable
about ten times more than \sonnet. A trained network is far cheaper per volume at inference. The difference
is the marginal cost of a \emph{new} concept or modality. For \ours, that cost is a new prompt. For a trained
model, it is a new labeled dataset and a new training run, about 9{,}200 A100 hours on 158 curated datasets for the
current state of the art~\cite{rokuss2025voxtell}. MLLM prices for a given capability also
fall~\cite{cottier2025llmprices}.

\makeatletter
\let\arxiv@ssect\cvprssect
\def\arxiv@llm{LLM Usage Statement}
\def\cvprssect#1{\def\arxiv@tmp{#1}\ifx\arxiv@tmp\arxiv@llm\expandafter\@firstoftwo\else\expandafter\@secondoftwo\fi
  {\endinput}{\arxiv@ssect{#1}}}
\section{Conclusions and Limitations} \label{sec:conclusion}
We presented \ours, a training-free pipeline in which a frozen, general-purpose MLLM segments 3D CT, MRI, and
PET volumes from text. Running a 2D MLLM segmenter on every slice fails. \ours instead lets the MLLM plan the target,
localize it from coarse to fine, and trace a few keyframes, while deterministic geometry assembles the outlines.
Without training or labeled data, \ours matches the trained state of the art on nine tumor and lesion benchmarks and
clearly leads on unseen concepts and modalities. Its errors are mostly recognition errors, which stronger backbones
reduce (\figLabel\ref{fig:backends}). As MLLMs improve and their prices fall, training-free segmentation becomes a
practical complement to dense 3D labels. \ours still trails trained segmenters on small lesions and on small or thin
organs. It is also slower than a single network pass and relies
on commercial APIs whose models change over time. Sending patient images to an external API must comply with local data-protection
rules. We release every prompt and model reply so all masks can be rebuilt without API calls.

\section*{LLM Usage Statement}
In this work, we used generative AI tools (specifically, Claude Code) to help write
the evaluation-harness code and to edit portions of this manuscript (experimental details and introduction).
The segmentation method itself \emph{uses} MLLMs as its vision backend, which is the subject of the paper and is fully described in
\secLabel\ref{sec:method}. All AI-assisted code was verified and tested for correctness by the authors, and all AI-drafted text was
reviewed and revised by the authors. We take full responsibility for the final content of this work.

\let\cvprssect\arxiv@ssect
\makeatother

\section*{Acknowledgements}
This work was supported by Anthropic AI for Science research grant.

{
    \small
    \bibliographystyle{ieeenat_fullname}
    \bibliography{egbib}
}

\clearpage
\setcounter{topnumber}{4}
\setcounter{dbltopnumber}{3}
\renewcommand{\topfraction}{0.9}
\renewcommand{\dbltopfraction}{0.95}
\renewcommand{\textfraction}{0.05}
\renewcommand{\floatpagefraction}{0.6}
\renewcommand{\dblfloatpagefraction}{0.6}
\appendix
\twocolumn[\centering\Large\textbf{Appendix}\vspace{1.0em}]

\lstdefinestyle{promptstyle}{basicstyle=\fontsize{7}{8}\selectfont\ttfamily, columns=fullflexible,
  keepspaces=true, breaklines=true, breakindent=0pt, showstringspaces=false, aboveskip=0pt, belowskip=0pt}
\newtcolorbox{promptbox}[1]{enhanced, breakable, colframe=black!55, colback=black!2, boxsep=1pt, left=3pt,
  right=3pt, top=2pt, bottom=2pt, fonttitle=\bfseries\small, title={#1}}

\section{Benchmarks, Protocol, and Implementation}
\label{app:benchmarks}
\begin{table*}[h]
\centering
\small
\setlength{\tabcolsep}{4pt}
\resizebox{\textwidth}{!}{%
\begin{tabular}{llllrrl}
\toprule
Group & Benchmark & Modality & Prompts & Cases (ours) & Cases~\cite{rokuss2025voxtell} & Notes \\
\midrule
Tumors and lesions & & & & & & \\
\quad unknown concept & FedBCa C2~\cite{bladder_cancer} & MRI (T2) & carcinoma & 47 & 47 & \\
 & MedSeg~\cite{medseg_database} & CT & esophageal cancer & 1 & 1 & \\
\quad cross-modality & QIN-Breast~\cite{qinbreast} & PET & fdg-avid breast cancer & 38 & 36 & earliest scan per patient \\
 & Sarcoma~\cite{vallieres2015sarcoma} & MRI (T1, T2) & soft tissue lesion & 102 & 102 & \\
 & PANTHER~\cite{betancourt_tarifa_2025_15192302} & MRI & pancreas tumor & \na & 92 & not accessible, not used \\
\quad unseen dataset & ACRIN-NSCLC~\cite{kinahan2019acrin-nsclc} & PET & fdg-avid tumor & 36 & 44 & radiologist-validated AIMI masks \\
 & ISBI MS~\cite{isbi-ms} & MRI (FLAIR) & MS lesion & 21 $\times$ 2 raters & 42 & \\
 & Adrenal-ACC~\cite{adrenalacc} & CT & tumor & 52 & 52 & \\
 & HCC-TACE~\cite{hcctace} & CT & hepatocellular carcinoma & 103 & 65 & superset; selection rule unpublished \\
 & BrainMetShare~\cite{stanfordBrainMets} & MRI (T1-gd) & lesion & 105 & 105 & \\
Anatomical & Pediatric-CT-SEG~\cite{pediatric-ct-seg} & CT & 23 structures & 21 & 21 & cases with all classes annotated \\
 & AeroPath~\cite{2023aeropath} & CT & lung, trachea & 27 & 27 & \\
 & VEELA~\cite{toprak2025veela} & CT & liver, hepatic / portal vessels & 20 & 20 & \\
 & HVSMR-2.0~\cite{pace2024hvsmr} & MRI & 8 heart structures & 60 & 60 & \\
 & PENGWIN~\cite{pengwin} & CT & sacrum, left / right hipbone & 100 & 100 & fragments merged per bone; under pathology in~\cite{rokuss2025voxtell} \\
 & RADCURE~\cite{radcure} & CT & head and neck & \na & 22 & controlled access, not used \\
3D semantic seg. & BTCV~\cite{BTCV} & CT & 13 organs (10 scored) & 30 & \na & in the training set of~\cite{rokuss2025voxtell} \\
 & AMOS-MRI~\cite{AMOS} & MRI & 15 organs (10 scored) & 60 & \na & its CT part is in that training set \\
 & BraTS 2023~\cite{Baid2021TheRB} & MRI (4 seq.) & WT, TC, ET & 100 of 1251 & \na & seeded subset \\
\bottomrule
\end{tabular}}
\caption{\textbf{Evaluation benchmarks}, grouped as in the main paper. Prompts are the label names used by the original
protocols~\cite{rokuss2025voxtell,Bourigault_2025_ICCV}. All volumes are read in a canonical LPS orientation, and DICOM
series and segmentation objects are converted once.}
\label{tab:benchmarks}
\end{table*}

Every benchmark is used only for evaluation. Multi-rater ground truth (ISBI) is scored per rater. For BTCV and AMOS,
``adrenal glands'' is scored as the union of the left and right predictions against the union of the ground truth, as in UKBOB.
For BraTS, the plan picks the MR sequence per region, using FLAIR for the whole tumor and contrast T1 for the core
and enhancing tumor.

\mysection{Benchmark groups}
We evaluate zero-shot on 17 public benchmarks with 923 volumes and 79 prompts. We follow the protocol
of~\cite{rokuss2025voxtell} on all of its held-out benchmarks that are publicly available and group them by target
type. The \emph{tumor and lesion} group comprises nine benchmarks that differ in how far they lie from the baselines'
training data~\cite{rokuss2025voxtell}. Two test unknown concepts, namely bladder carcinoma in T2 MRI (FedBCa
center 2~\cite{bladder_cancer}) and an esophageal tumor in CT (MedSeg~\cite{medseg_database}, one volume). Two test
known concepts in a new modality, namely breast cancer in PET (QIN-Breast~\cite{qinbreast}) and soft-tissue sarcoma in
T1 and T2 MRI~\cite{vallieres2015sarcoma}. Five are unseen datasets of known concepts. They cover lung tumors in
FDG-PET (ACRIN-NSCLC~\cite{kinahan2019acrin-nsclc}), multiple sclerosis lesions in FLAIR (ISBI~\cite{isbi-ms}, two
raters), adrenal tumors (Adrenal-ACC~\cite{adrenalacc}), hepatocellular carcinoma (HCC-TACE~\cite{hcctace}), and brain
metastases (BrainMetShare~\cite{stanfordBrainMets}). The \emph{anatomical} group has five benchmarks. They cover 23
structures in pediatric CT (Pediatric-CT-SEG~\cite{pediatric-ct-seg}), lung and airway (AeroPath~\cite{2023aeropath}),
liver and vessels (VEELA~\cite{toprak2025veela}), the whole heart in MRI (HVSMR-2.0~\cite{pace2024hvsmr}, eight
structures), and the pelvic bones of patients with fractures (PENGWIN~\cite{pengwin}). PENGWIN is listed under
pathology in~\cite{rokuss2025voxtell}, but its targets are bones. We add three standard benchmarks of 3D \emph{semantic
segmentation}, on which trained models are evaluated in their own domain~\cite{Bourigault_2025_ICCV}. These cover the
ten abdominal organs shared by BTCV~\cite{BTCV} (CT) and AMOS~\cite{AMOS} (MRI), as well as brain tumors in a seeded
100-case subset of BraTS~2023~\cite{Baid2021TheRB}. RADCURE and PANTHER could not be obtained and are
left out of all means.

\mysection{Protocol and baselines}
Every method receives the benchmark's label name as its prompt, with no prompt tuning and no dataset-specific settings.
We report Dice $\times 100$, averaged over cases per prompt and then over prompts per benchmark. We compare with the
trained text-promptable models evaluated in~\cite{rokuss2025voxtell} and quote their published numbers. For the state
of the art, VoxTell, we run the released checkpoint ourselves on our exact cases, ground truth, and metric. Our run
stays within 3.5 Dice of its published numbers wherever the case sets match
(\secLabel\ref{app:results}). On three benchmarks, the published case selection is not available, and ours differs
(ACRIN 36 vs.\ 44 scans, HCC-TACE 103 vs.\ 65, and QIN 38 vs.\ 36). In those columns, only the VoxTell row is paired with
ours. Every baseline was trained for medical segmentation on voxel labels. \ours adds no training, and its backbone is
a general-purpose multimodal large language model (MLLM) that reads
images and writes text. The training data of such MLLMs is not public and may include
medical images. By zero-shot, we therefore mean that \ours uses no label, example, or tuning from any benchmark. The
per-organ and per-region tables are in \secLabel\ref{app:complete}.

\mysection{Backbones and implementation}
We run \ours with three backbones from one model family through the provider API, using identical prompts and
settings. They are \haiku (small), \sonnet (mid-size)~\cite{anthropic2026claude55}, and \fable (strongest). No
other component is learned or tuned per benchmark. \fable costs about ten times more per volume than \sonnet
(\secLabel\ref{app:cost}). We therefore run it on all nine tumor and lesion benchmarks (505 volumes) and on BraTS, and
we run the two smaller backbones on every benchmark. The anatomical benchmarks, whose volumes hold up to 23 targets
each, and the multi-organ benchmarks BTCV and AMOS are reported with the two smaller backbones only, since running
\fable on them is cost-prohibitive. Beyond this family, we also evaluate the unchanged pipeline with 25 backbones from six
providers on ten pelvic CT volumes (\secLabel\ref{app:backends}). In the main paper, \ours without a backbone name refers to the strongest backbone in
the corresponding table. The qualitative figures show \fable on tumors and lesions, and \sonnet on anatomy. Plan and
localization calls use medium reasoning effort, and contour calls use the highest available effort. All settings
(\secLabel\ref{app:method}) were chosen on development cases of the anatomical benchmarks and confirmed on held-out
cases of the same benchmarks.
The calls for a volume
run in parallel, so a volume with one target takes a few minutes, and a multi-organ scan takes tens of minutes.

\section{Method Details}
\label{app:method}

\mysection{Notation}
We use the notation of the main paper's Method. The volume is $\mathbf{V}$, the prompt is $\mathbf{q}$, and the frozen MLLM is
$\mathbf{F}$. The coordinate $z$ is the position of an axial slice along the body axis, and $\mathbf{u}=(u,v)$ is a pixel within the
slice. Fine localization gives one box $\mathbf{B}_z$ per slice and the extent $[z_{\min},z_{\max}]$ of the target.
The keyframes are the slices $z_k$. On keyframe $z_k$, the MLLM returns loops of points $\mathbf{p}_j$, the curve
$\mathbf{C}_k$ of \eqn{eq:spline} encloses the region $\Omega_k$, and $\mathbf{m}_k$ is the keyframe mask. The signed
distance maps $\boldsymbol{\phi}_k$ (\eqn{eq:dist}) are blended into the assembled mask $\widetilde{\mathbf{M}}$
(\eqn{eq:sdf}), and intensity snapping (\eqn{eq:snap}) gives the final mask $\mathbf{M}$.

\mysection{Settings}
One set of values is used for every benchmark, modality, and backbone. The plan gives the expected size $s$ of the
target in mm. Coarse localization shows $N$ axial views spaced at most $s/2$ apart, with $16\le N\le 48$, plus one
coronal and one sagittal view. Fine localization shows full axial slices 8~mm apart (at most 24 per call) over the
coarse extent padded by one coarse spacing, and it returns the boxes $\mathbf{B}_z$ that set $[z_{\min},z_{\max}]$.
Keyframes $z_k$ are placed in this extent at a spacing of $g=\min(8\,\mathrm{mm}, s/2)/3$, with one probe beyond each end.
Each contour call holds at most 8 keyframes. Views are rendered at 768 pixels on their long side, and the zoomed crops around
$\mathbf{B}_{z_k}$ at 1024 pixels. A loop has 8 to 40 points $\mathbf{p}_j$. A trace that touches the
crop border is re-traced on an enlarged crop. When an end keyframe or a probe is marked ``present'', the extent grows
by one more probe, up to three times. The half-width $b$ of the snapping band is 0.08 times the median
cross-sectional diameter of the mask, clipped to $[1, 8]$~mm. A case with several prompts shares the plan call, and targets planned in the same display window share the
coarse call. From fine localization on, the targets are handled independently and in parallel. For multi-sequence MR,
the plan also picks the sequence per target.

\mysection{Rendering and coordinates}
Slices are resampled to square physical pixels and rendered at no more than 1536 pixels on the long side. Beyond that
size, the API downsamples images, and the returned coordinates degrade. On synthetic shapes, the three backbones of the
main results return boxes with an IoU of 0.95 to 0.99 against the true boxes at 448 to 1536 pixels. Replies are parsed by a
tolerant JSON reader. Boxes and points are then mapped from image pixels back to the pixel $\mathbf{u}=(u,v)$ of the
slice at position $z$ through each view's crop and scale. Backbones that answer on a normalized grid instead of in
pixels receive prompts with coordinate instructions and JSON formats rewritten for that grid. Their replies are
mapped back in the same way (\secLabel\ref{app:backends}).

\mysection{Multi-part targets}
When the plan says ``multiple'', the fine stage scans the whole volume in up to three calls of 24 slices. Boxes are
linked across slices by overlap into tracks. Each track is contoured with at most 6 keyframes, up to a total of 64
keyframes per target, prioritizing the largest tracks first. The track masks are assembled separately and then merged.

\mysection{Snapping}
This paragraph completes \eqn{eq:snap}. The inside statistics $\mu_{\mathrm{in}}$ and $\sigma_{\mathrm{in}}$ are
the median and the scaled median absolute deviation of the intensities of $\mathbf{V}$ in the core of the assembled
mask $\widetilde{\mathbf{M}}$, defined as the interior voxels deeper than $b$ ($d(\mathbf{u},z)>b$). For a thin target, the core is the inner half of the mask. The outside statistics
$\mu_{\mathrm{out}}$ and $\sigma_{\mathrm{out}}$ are measured in a shell 6~mm wide that starts at distance $b$ outside
the mask. Snapping is applied only if the two sides are separable, meaning
$\lvert\mu_{\mathrm{in}}-\mu_{\mathrm{out}}\rvert\,/\sqrt{(\sigma_{\mathrm{in}}^2+\sigma_{\mathrm{out}}^2)/2}\ge 0.5$.
The cost $\ell_c$ is the negative log-likelihood of a Laplace distribution with location $\mu_c$ and scale $\sigma_c$,
up to a constant. After the assignment, the mask is smoothed on each slice by an opening and a closing with a
$3\times3$ element, and connected components that hold no core voxel are discarded.

\section{Discussion of the Main Results}
\label{app:discussion}
This section describes the main tables and figures. The per-organ and per-region tables are in
\secLabel\ref{app:complete}, the prompt wording study in \secLabel\ref{app:wording}, and the slice-by-slice baseline in
\secLabel\ref{app:llmseg_slice}.

\mysection{Tumors and lesions}
\tblLabel\ref{tab:main} compares \ours with three backbones against trained text-promptable models on the nine tumor and
lesion benchmarks. The columns are grouped into unknown concepts, known concepts in a new modality, and unseen datasets
of known concepts.

\mysection{Anatomical structures}
\tblLabel\ref{tab:anatomy} reports results on the five anatomical benchmarks for the two smaller backbones.

\mysection{3D semantic segmentation}
\tblLabel\ref{tab:standard} places \ours among models trained with voxel labels of the same structures. Per-organ and
per-region scores, including the strongest backbone on BraTS, are in \tblLabel\ref{tab:organs} and
\tblLabel\ref{tab:brats}.

\mysection{Qualitative results}
\figLabel\ref{fig:qual_anat} and \figLabel\ref{fig:qual_path} show representative cases, and
\figLabel\ref{fig:qual_best} and \figLabel\ref{fig:failures} add best and failure cases. The tumor and lesion rows
show \fable, and the anatomy rows show \sonnet.

\section{Per-Organ and Per-Region Results}
\label{app:complete}
The main paper groups the benchmarks by target type and reports one row per method. This section gives the
per-organ and per-region scores in the layouts of the original protocol~\cite{Bourigault_2025_ICCV}, with the
published numbers of trained models and the variants of our pipeline in which MedSAM2~\cite{MedSAM2} replaces the MLLM
tracer.

\mysection{MedSAM2 in place of the MLLM tracer}
The MedSAM2 rows of \tblLabel\ref{tab:brats} use the boxes of our fine localization, the last MLLM stage before tracing.
They are therefore ablations of our pipeline and not independent baselines.

\begin{table*}[t]
\centering
\setlength{\tabcolsep}{3.2pt}
\renewcommand{\arraystretch}{0.9}
\begin{adjustbox}{max width=\textwidth}
\begin{tabular}{ll|cccccccccc|c}
\toprule
Dataset & Method & Spleen & R.\ Kid. & L.\ Kid. & Gall. & Eso. & Liver & Stom. & IVC & AG & Aorta & Mean \\
\midrule
\multirow{7}{*}{\shortstack{\textbf{BTCV}\\\CT}} & TotalVibeSeg.~\cite{graf2024totalvibesegmentator} & 84.8 & 72.1 & 80.1 & 78.5 & 79.7 & 79.5 & 73.7 & 71.5 & \textbf{86.1} & \second{86.2} & 79.2 \\
 & Swin-BOB~\cite{Bourigault_2025_ICCV} & \second{88.3} & \second{88.4} & 93.2 & 79.5 & 79.0 & \textbf{94.6} & \second{88.5} & \second{87.1} & 78.4 & 79.9 & 85.6 \\
 & Swin-BOB + vol.\ filter & 88.1 & 87.0 & \second{93.9} & \second{81.2} & \second{80.1} & \second{92.5} & 87.3 & 85.8 & 78.1 & 85.2 & \second{87.5} \\
 & Swin-BOB + SOLF & \textbf{89.1} & \textbf{89.0} & \textbf{94.9} & \textbf{82.3} & \textbf{88.1} & 89.7 & \textbf{89.9} & \textbf{89.1} & \second{82.4} & \textbf{87.1} & \textbf{88.2} \\
\cmidrule{2-13}
 & VoxTell$^{\ast\dagger}$~\cite{rokuss2025voxtell} & \textbf{95.6} & \textbf{91.5} & \textbf{93.7} & \textbf{81.8} & \textbf{78.7} & \textbf{96.5} & \textbf{90.7} & \textbf{87.0} & \textbf{74.8} & \textbf{92.0} & \textbf{88.2} \\
\oursrow  & \oursh & 91.4 & 86.9 & 85.7 & 56.8 & 52.6 & 92.0 & 77.7 & 65.1 & 26.3 & 83.7 & 71.8 \\
\oursrow  & \ourss & \second{91.7} & \second{90.6} & \second{91.1} & \second{58.6} & \second{62.0} & \second{92.5} & \second{82.1} & \second{69.0} & \second{42.4} & \second{87.4} & \second{76.7} \\
\midrule
\multirow{7}{*}{\shortstack{\textbf{AMOS}\\\MRI}} & TotalVibeSeg.~\cite{graf2024totalvibesegmentator} & 82.3 & 69.7 & 81.4 & \textbf{78.2} & \textbf{78.6} & 80.2 & 75.9 & 73.8 & \textbf{87.9} & 88.1 & 79.6 \\
 & Swin-BOB~\cite{Bourigault_2025_ICCV} & 90.8 & 93.1 & 94.2 & 65.7 & 65.8 & 95.8 & 82.2 & \second{87.4} & 52.9 & 90.6 & 81.8 \\
 & Swin-BOB + vol.\ filter & \second{91.0} & \second{94.0} & \second{95.1} & 65.8 & 66.7 & \second{96.6} & \second{83.2} & \textbf{88.2} & 62.1 & \second{91.8} & \second{83.2} \\
 & Swin-BOB + SOLF & \textbf{91.9} & \textbf{94.3} & \textbf{96.2} & \second{66.4} & \second{67.2} & \textbf{96.9} & \textbf{83.8} & \textbf{88.2} & \second{63.1} & \textbf{92.4} & \textbf{84.0} \\
\cmidrule{2-13}
 & VoxTell$^{\ast\dagger}$~\cite{rokuss2025voxtell} & \textbf{95.4} & \textbf{95.6} & \textbf{95.4} & \textbf{82.7} & \textbf{66.8} & \textbf{96.5} & \textbf{85.0} & \textbf{84.9} & \textbf{69.8} & \textbf{91.7} & \textbf{86.4} \\
\oursrow  & \oursh & 87.4 & 90.3 & 89.8 & 55.9 & 23.9 & 89.7 & 50.7 & 41.6 & 18.0 & 79.9 & 62.7 \\
\oursrow  & \ourss & \second{91.2} & \second{91.7} & \second{91.9} & \second{65.9} & \second{40.0} & \second{90.5} & \second{63.9} & \second{50.8} & \second{28.4} & \second{84.4} & \second{69.9} \\
 
\bottomrule
\end{tabular}
\end{adjustbox}
\caption{\textbf{Abdominal organ segmentation.} We show the Dice scores on BTCV (CT) and AMOS (MRI) for the ten organs the two datasets
share (AG: left and right adrenal glands). The top of each block holds published zero-shot numbers of models trained on
UK Biobank MRI labels~\cite{Bourigault_2025_ICCV,graf2024totalvibesegmentator}. The bottom holds our evaluation on all
labeled cases (BTCV 30, AMOS-MRI 60). $^\ast$The released checkpoint run by us. $^\dagger$It was trained on BTCV
and AMOS CT. \textbf{Bold}: best,
\underline{underline}: second best per block.}
\label{tab:organs}
\end{table*}

\begin{table}[t]
\centering
\setlength{\tabcolsep}{4pt}
\renewcommand{\arraystretch}{0.86}
\begin{adjustbox}{max width=\columnwidth}
\begin{tabular}{l|ccc|c}
\toprule
Method & WT & TC & ET & Mean \\
\midrule
\multicolumn{5}{l}{\textit{Supervised on BraTS (UKBOB paper)}} \\
UNet~\cite{ronneberger2015unet} & \na & \na & \na & 54.4 \\
V-Net & \na & \na & \na & 84.2 \\
ResUNet++ & \na & \na & \na & 78.4 \\
Attention U-Net & \na & \na & \na & 79.8 \\
nnFormer & \na & \na & \na & 81.2 \\
UNETR & \na & \na & \na & 87.1 \\
SegResNet & \na & \na & \na & \second{89.0} \\
TotalVibeSeg. & \na & \na & \na & 83.0 \\
Swin-UNETR & \na & \na & \na & 88.6 \\
Swin-BOB & \na & \na & \na & \textbf{89.4} \\
\midrule
\multicolumn{5}{l}{\textit{Our evaluation (identical 100 cases)}} \\
VoxTell$^{\ast\dagger}$~\cite{rokuss2025voxtell} & \textbf{90.9} & \textbf{90.9} & \textbf{83.4} & \textbf{88.4} \\
MedSAM2, \sonnet boxes & 84.6 & 50.8 & 16.2 & 50.6 \\
MedSAM2, \fable boxes & \second{88.4} & 53.6 & 17.4 & 53.1 \\
\oursrow \oursh & 79.3 & 51.9 & 55.2 & 62.2 \\
\oursrow \ourss & 83.1 & 73.3 & 65.4 & 73.9 \\
\oursrow \oursf & 85.8 & \second{76.4} & \second{69.0} & \second{77.1} \\
 
\bottomrule
\end{tabular}
\end{adjustbox}
\caption{\textbf{Brain tumor segmentation.} We show the Dice scores on BraTS 2023 (MRI) for the whole tumor (WT),
tumor core (TC), and enhancing tumor (ET). The published models are trained on BraTS, and their numbers are those
reported on its test split~\cite{Bourigault_2025_ICCV}. Our evaluation uses a seeded 100-case subset. $^\ast$The
released checkpoint run by us. $^\dagger$It was trained on BraTS 2024. The MedSAM2 rows are variants of our
pipeline in which MedSAM2 replaces the MLLM tracer and is prompted by the named backbone's boxes. \ours is zero-shot.
\textbf{Bold}: best, \underline{underline}: second best per block.}
\label{tab:brats}
\end{table}

\mysection{Abdominal organs}
\tblLabel\ref{tab:organs} gives the per-organ scores on BTCV and AMOS. On these two benchmarks, the plan's shape field
counts a hollow organ such as the stomach as solid and reserves ``shell'' for targets that are themselves thin layers,
such as skin.

\section{Additional Results}
\label{app:results}

\mysection{Reproducing the trained baseline}
We run the checkpoint released with~\cite{rokuss2025voxtell} (marked $^\ast$ in the tables of \secLabel\ref{app:complete}). A later release of the
same model was also trained on the test sets used here and is therefore not a zero-shot baseline. Our run matches the
published numbers within $-3.4$ to $+0.0$ Dice on the eleven benchmarks with identical case sets, and within one point on
seven of them. It differs on the three benchmarks whose case sets are not the same. On ACRIN, it reaches 62.3 instead of
83.2 (36 vs.\ 44 scans), because our set holds small tumors that it misses. On HCC-TACE, it reaches 66.3 instead of 73.2
(103 vs.\ 65 scans), and on QIN, it reaches 69.7 instead of 72.3 (38 vs.\ 36 scans). We use our run for every paired comparison.

\begin{figure*}[t]
  \centering
  \setlength{\qgw}{0.128\textwidth}
  \setlength{\tabcolsep}{0.6pt}
  \renewcommand{\arraystretch}{0.4}
  \begin{tabular}{@{}c*{5}{c}@{\hspace{3pt}}*{2}{c}@{}}
   & \qhd{Image} & \qhd{Ground truth} & \qhd{\ours (ours)} & \qhd{VoxTell} & \qhd{MLLM boxes + MedSAM2} & \qhd{Ground truth (3D)} & \qhd{\ours (ours, 3D)} \\
  \qrow{Adrenal-ACC (CT)}{tumor} & \qimh{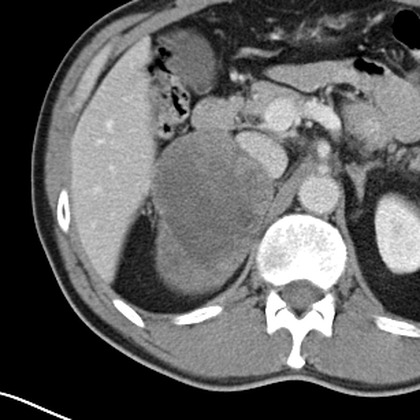} & \qimh{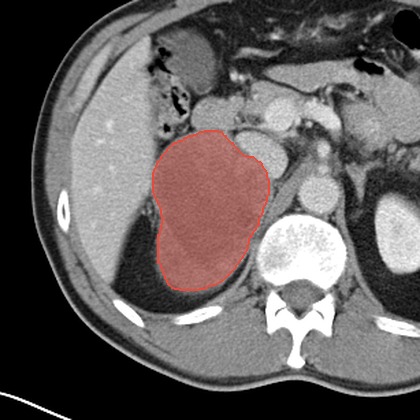} & \qimh{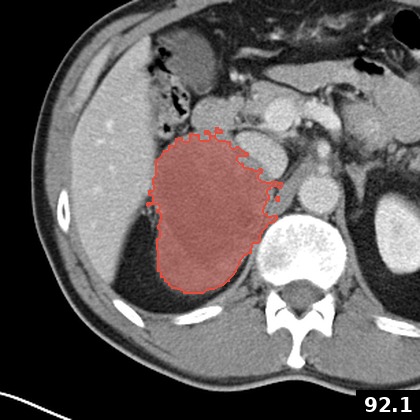} & \qimh{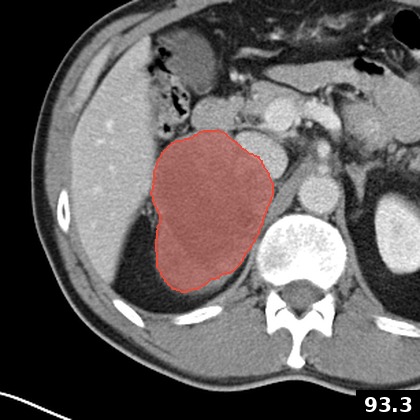} & \qimh{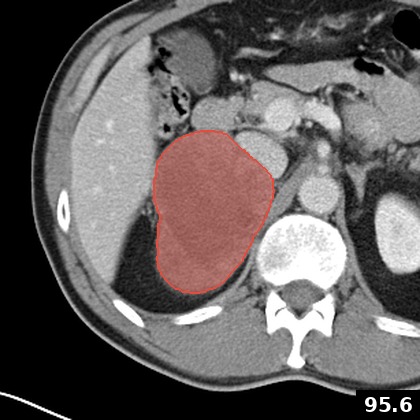} & \qimh{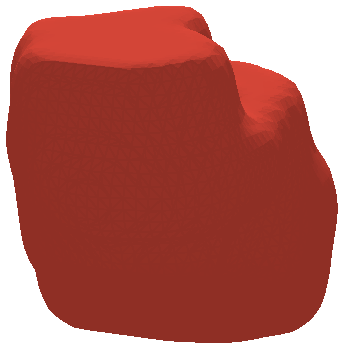} & \qimh{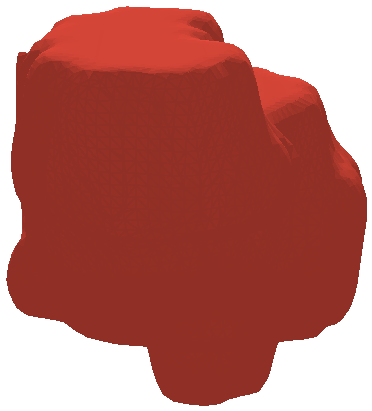} \\
  \qrow{ACRIN (PET)}{fdg-avid tumor} & \qimh{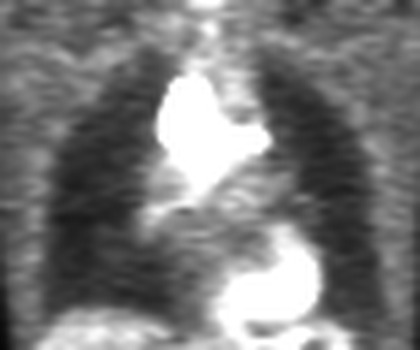} & \qimh{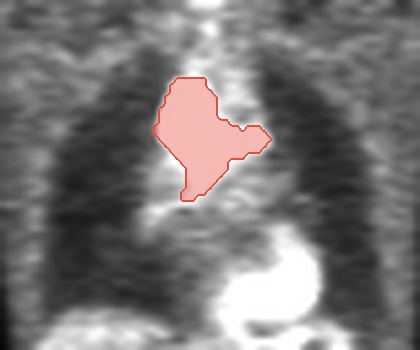} & \qimh{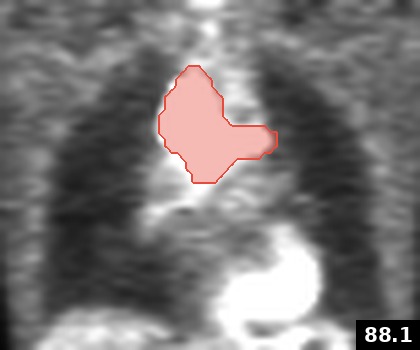} & \qimh{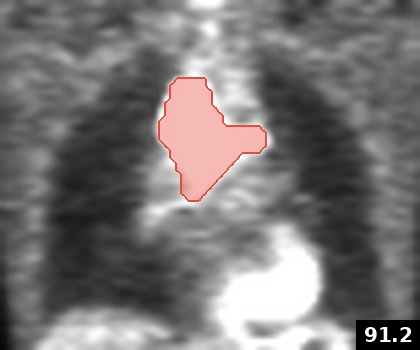} & \qimh{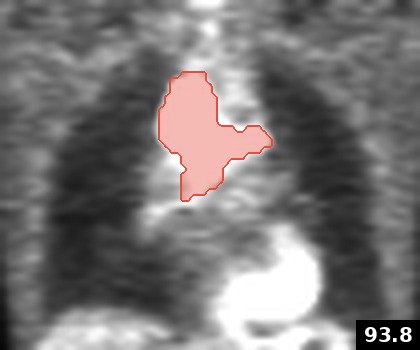} & \qimh{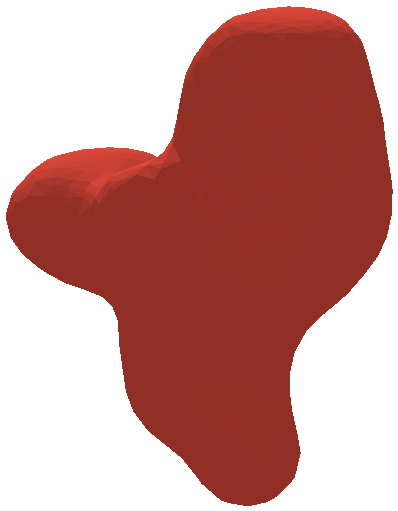} & \qimh{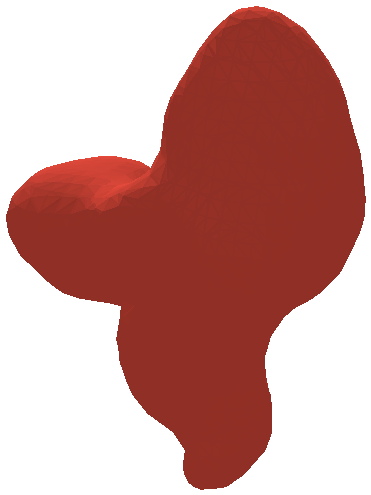} \\
  \qrow{BrainMetShare}{lesion} & \qimh{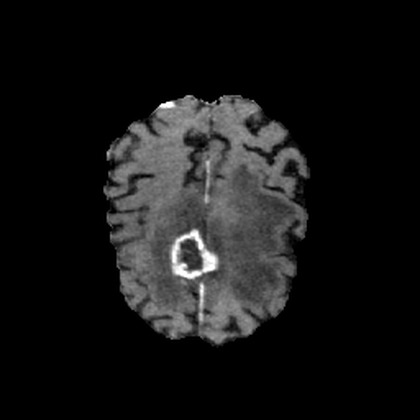} & \qimh{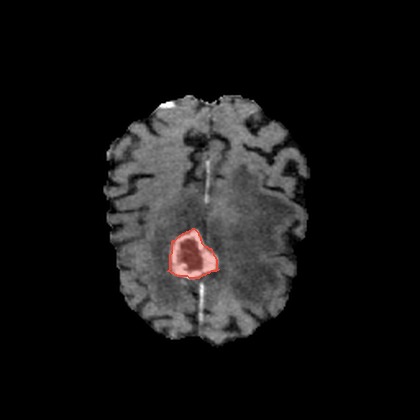} & \qimh{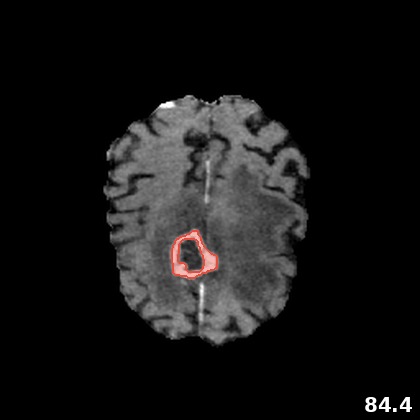} & \qimh{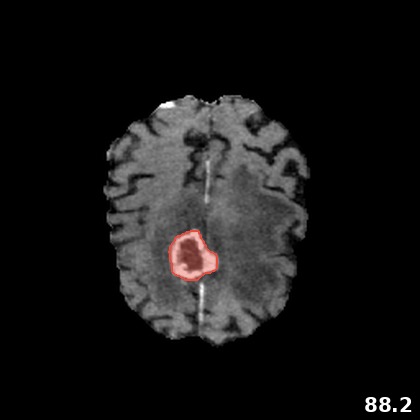} & \qimh{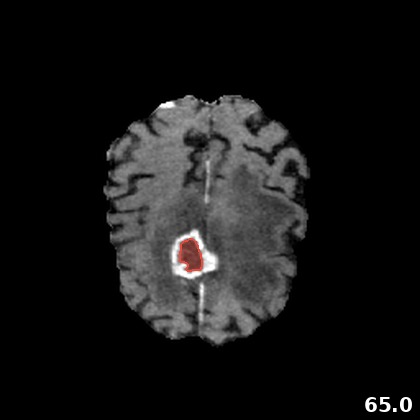} & \qimh{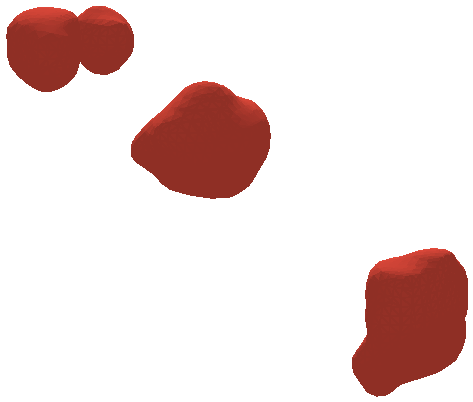} & \qimh{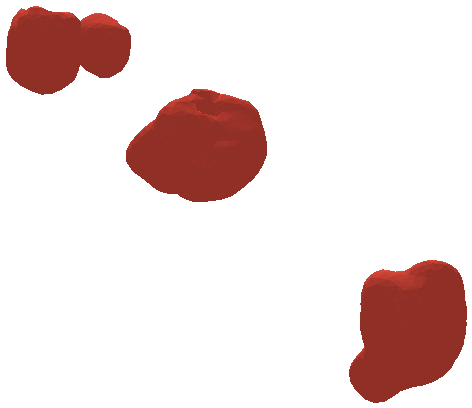} \\
  \qrow{Sarcoma (T2)}{soft tissue lesion} & \qimh{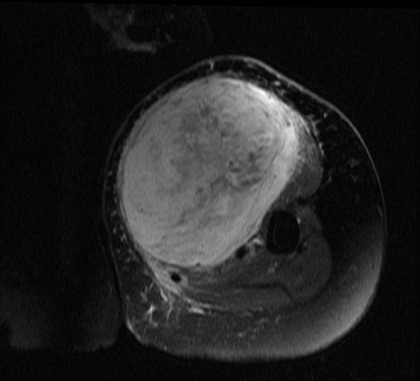} & \qimh{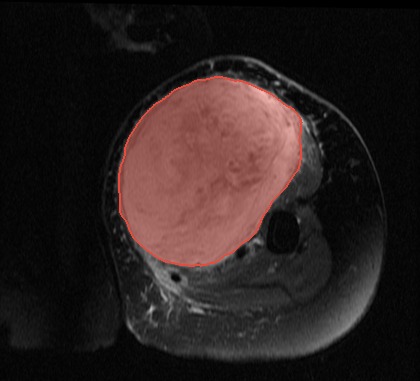} & \qimh{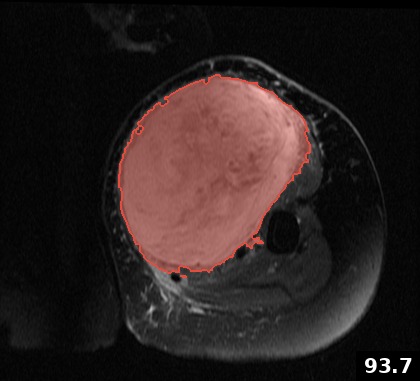} & \qimh{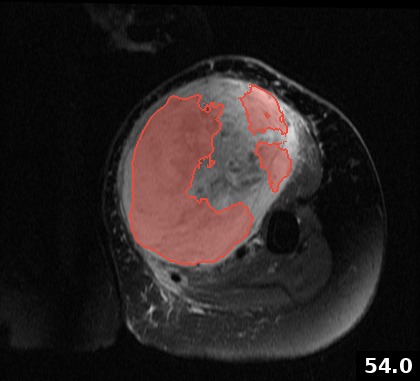} & \qimh{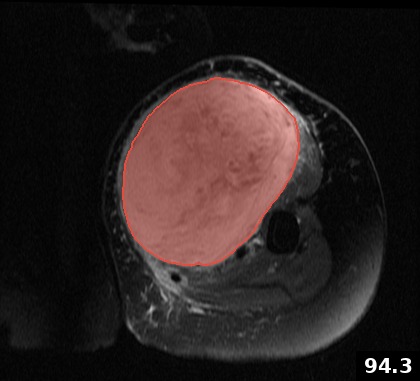} & \qimh{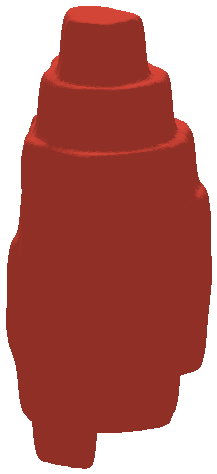} & \qimh{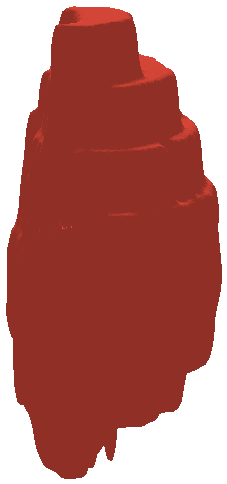} \\
  \qrow{FedBCa (T2)}{carcinoma} & \qimh{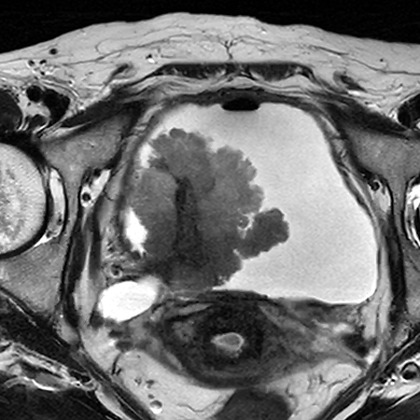} & \qimh{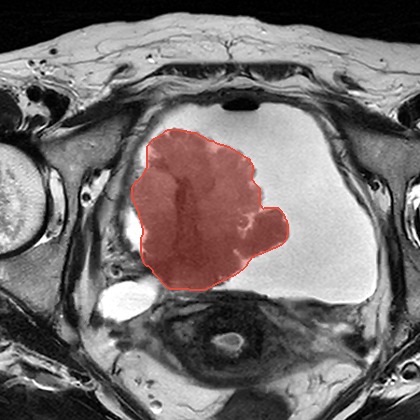} & \qimh{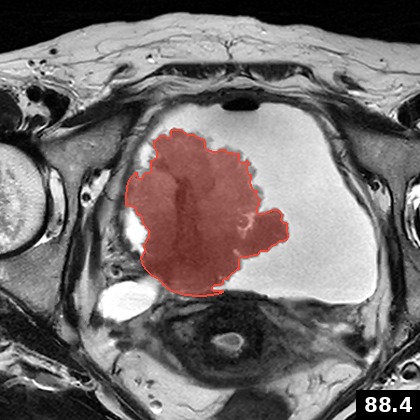} & \qimh{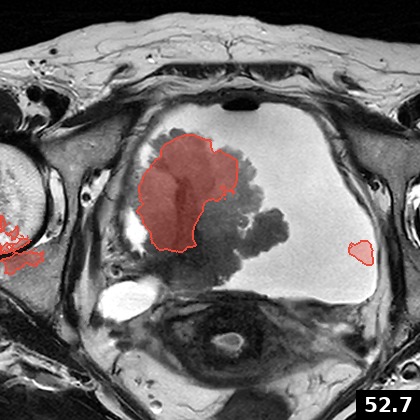} & \qimh{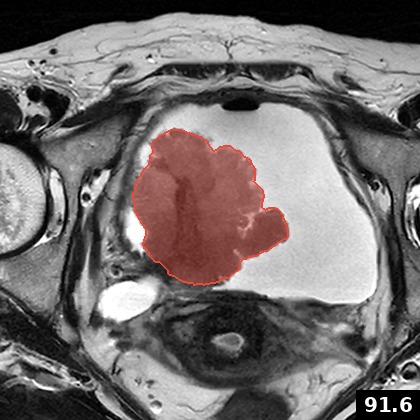} & \qimh{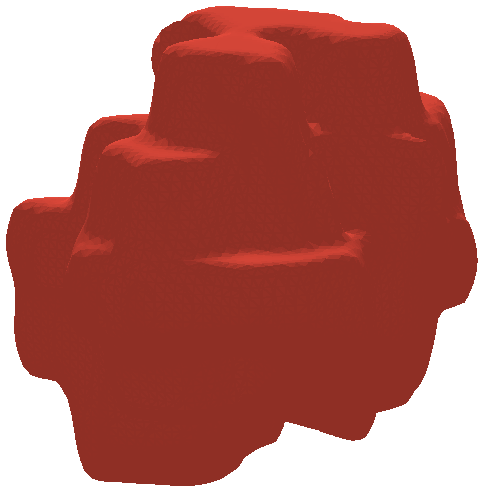} & \qimh{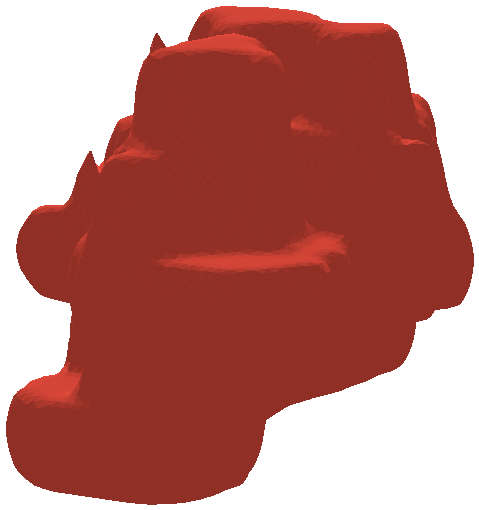} \\
  \qrow{HCC-TACE (CT)}{hepatocell. carcinoma} & \qimh{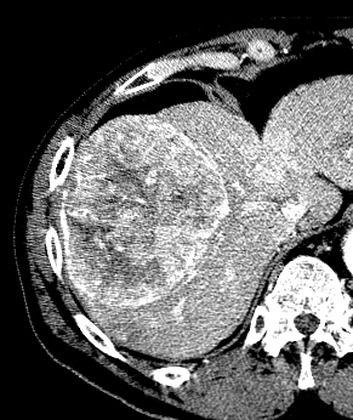} & \qimh{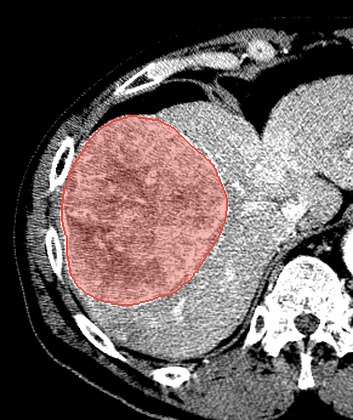} & \qimh{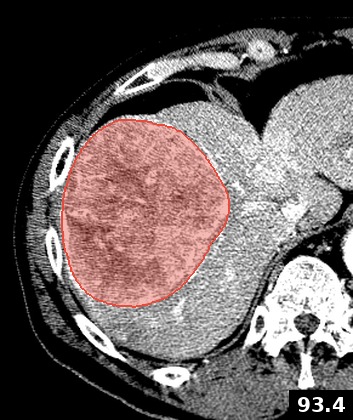} & \qimh{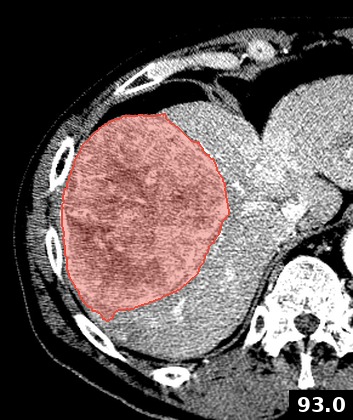} & \qimh{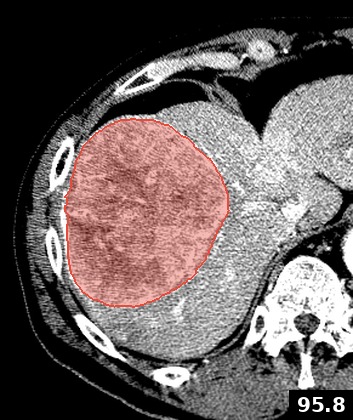} & \qimh{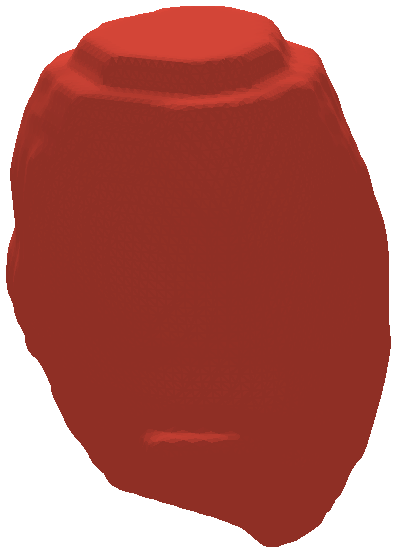} & \qimh{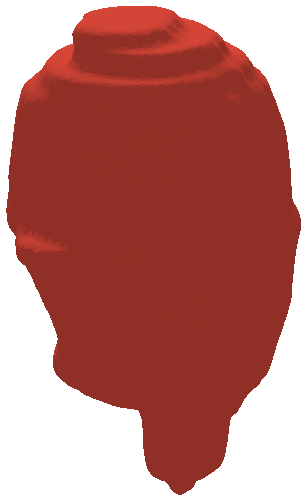} \\
  \qrow{ISBI (FLAIR)}{MS lesion} & \qimh{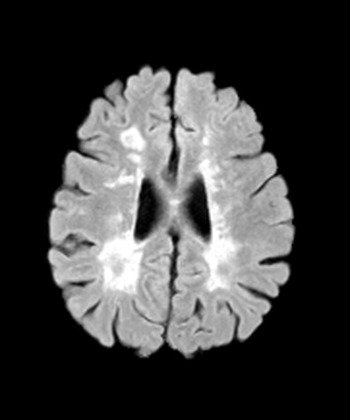} & \qimh{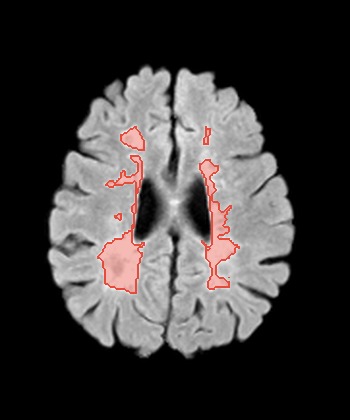} & \qimh{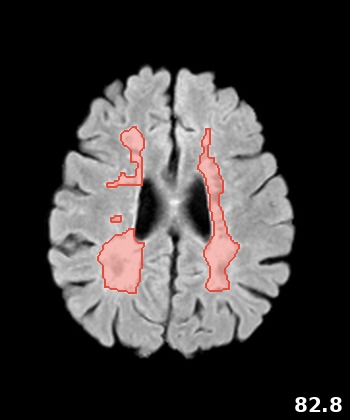} & \qimh{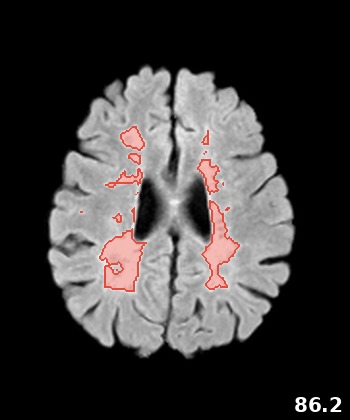} & \qimh{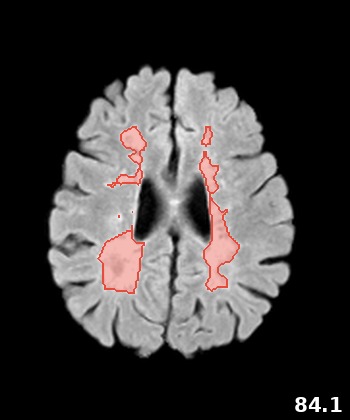} & \qimh{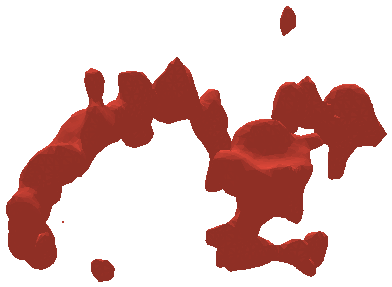} & \qimh{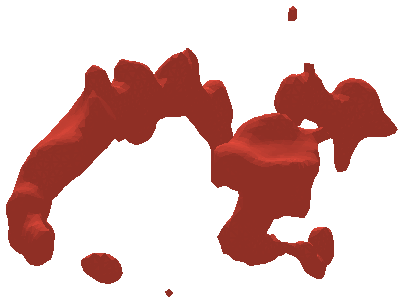} \\
  \end{tabular}
  \caption{\textbf{Best case per benchmark.} Each row shows the highest-scoring case of \ours among the ten volumes with
  stored masks, alongside VoxTell~\cite{rokuss2025voxtell} and MedSAM2~\cite{MedSAM2} prompted by the same MLLM's boxes.
  \ours uses \fable here. Each chip reports the Dice score of the shown targets.}
  \label{fig:qual_best}
\end{figure*}

\mysection{Qualitative results, best cases}
Masks were stored for the first ten volumes of each benchmark, and the qualitative figures select from them.
\figLabel\ref{fig:qual_best} shows the best such case of \ours (\fable) on each tumor and lesion benchmark, next to VoxTell and the
variant of our pipeline in which MedSAM2 replaces the MLLM tracer.

\begin{figure*}[t]
  \centering
  \setlength{\qgw}{0.17\textwidth}
  \setlength{\tabcolsep}{0.6pt}
  \renewcommand{\arraystretch}{0.4}
  \begin{tabular}{@{}c*{5}{c}@{}}
   & \qhd{Image} & \qhd{Ground truth} & \qhd{\ours (ours)} & \qhd{VoxTell} & \qhd{MLLM boxes + MedSAM2} \\
  \qrow{BTCV (CT)}{13 organs} & \qimh{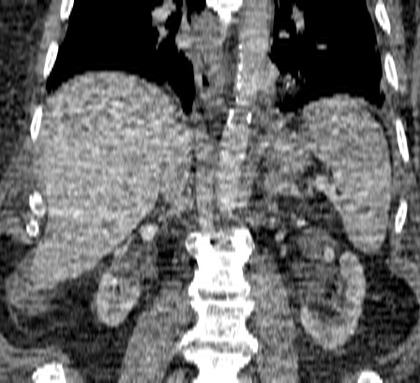} & \qimh{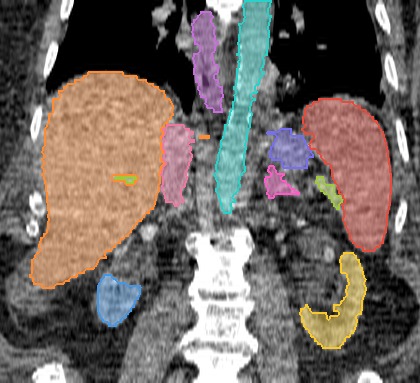} & \qimh{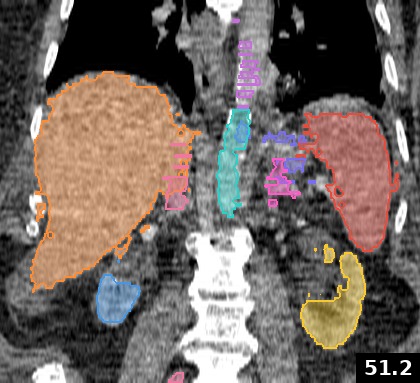} & \qimh{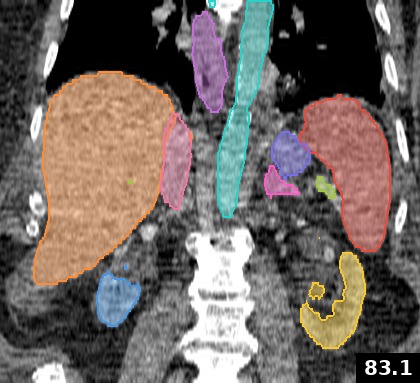} & \qimh{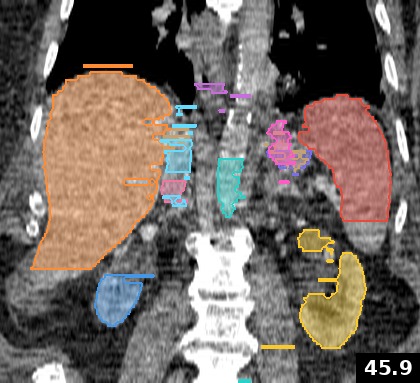} \\
  \qrow{HVSMR (MRI)}{8 heart structures} & \qimh{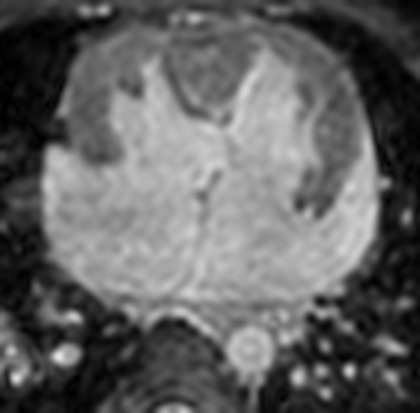} & \qimh{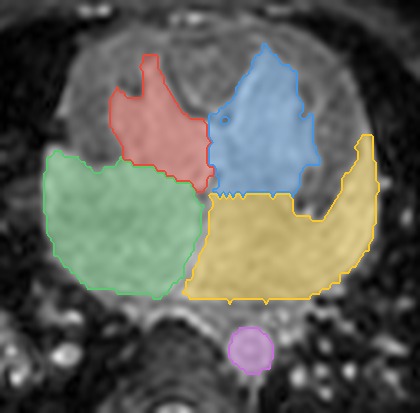} & \qimh{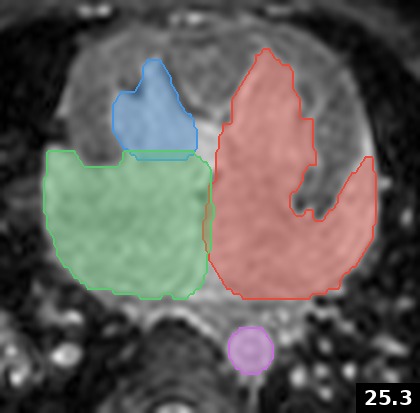} & \qimh{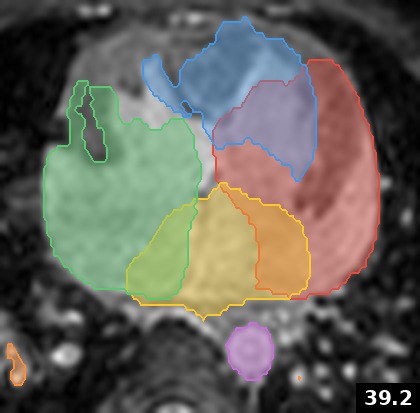} & \qimh{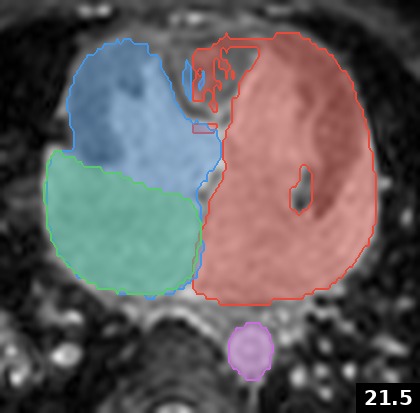} \\
  \qrow{HCC-TACE (CT)}{hepatocell. carcinoma} & \qimh{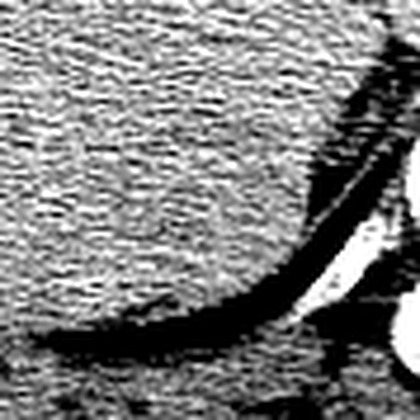} & \qimh{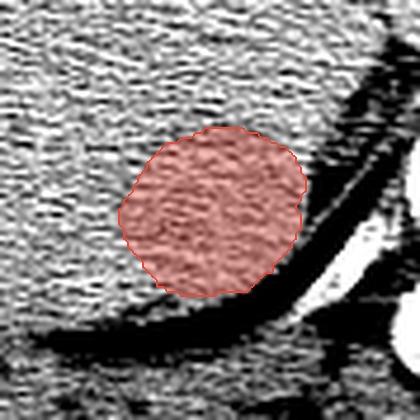} & \qimh{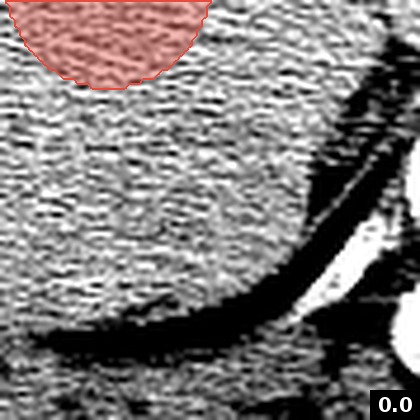} & \qimh{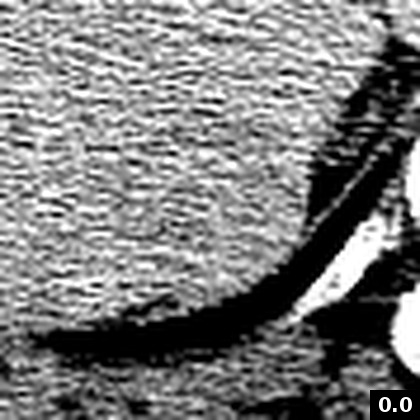} & \qimh{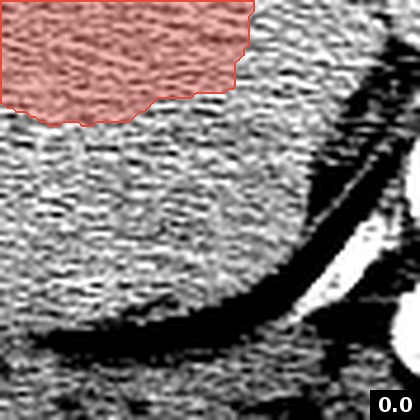} \\
  \qrow{ACRIN (PET)}{fdg-avid tumor} & \qimh{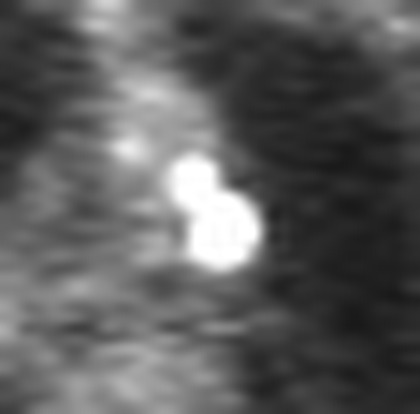} & \qimh{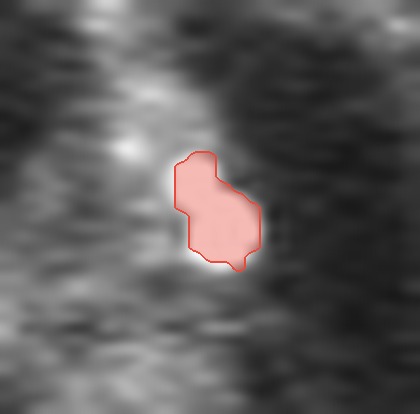} & \qimh{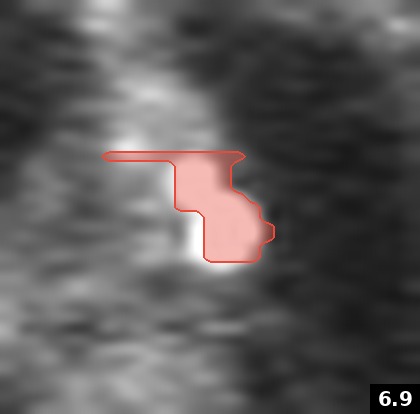} & \qimh{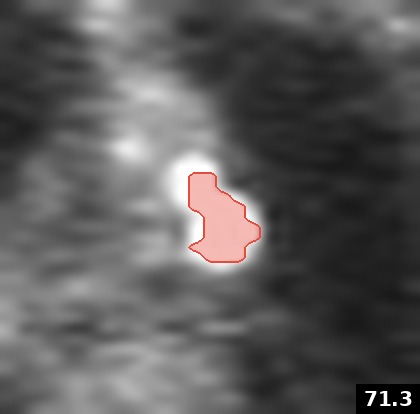} & \qimh{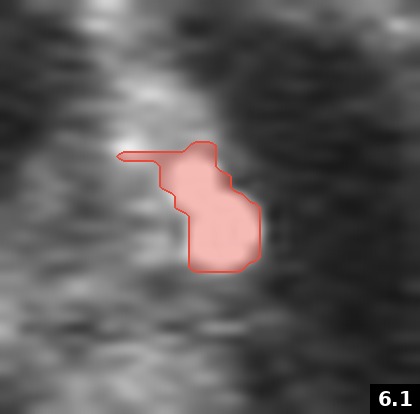} \\
  \qrow{BrainMetShare}{lesion} & \qimh{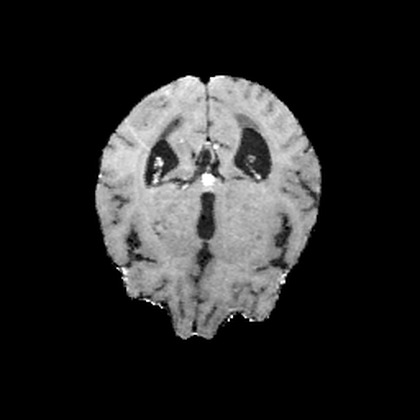} & \qimh{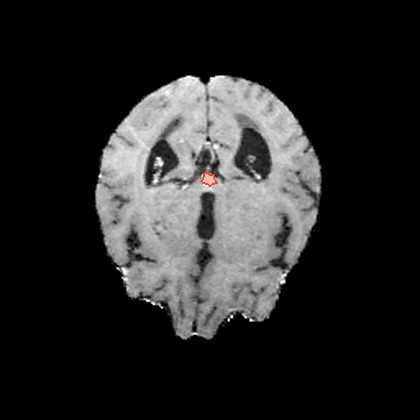} & \qimh{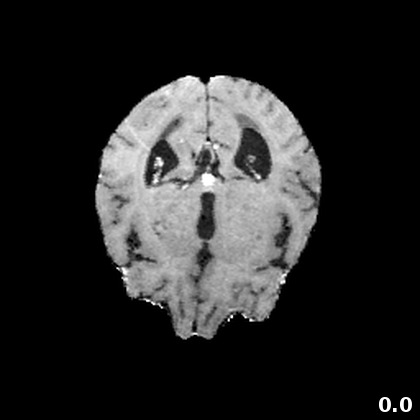} & \qimh{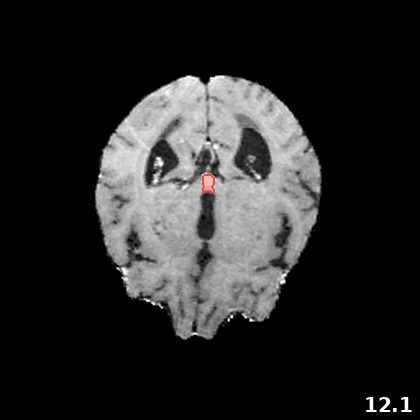} & \qimh{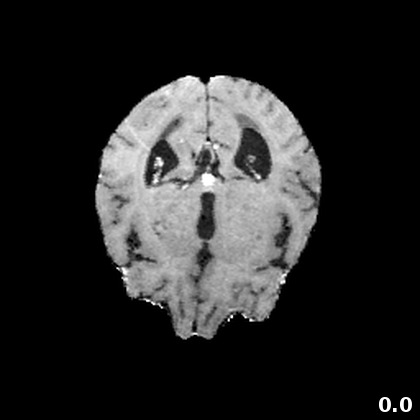} \\
  \end{tabular}
  \caption{\textbf{Failure cases.} Each row shows the lowest-scoring case of \ours on a benchmark among the ten
  volumes with stored masks, alongside VoxTell and MedSAM2 prompted by our MLLM's boxes. The tumor and lesion rows
  show \fable for both. The two anatomy rows show \sonnet for \ours (since \fable was not run on anatomy) and MedSAM2
  prompted by \haiku's boxes. Most are recognition errors. Thin abdominal
  organs (esophagus, adrenal glands) are missed, and heart chambers are labeled as their neighbors. A liver region is
  traced instead of the tumor. A lung tumor in PET is found, but the mask spreads over uptake far beyond it. A small
  metastasis is missed.}
  \label{fig:failures}
\end{figure*}

\mysection{Failure cases}
\figLabel\ref{fig:failures} shows the lowest-scoring of these cases. Most of them are recognition errors.

\subsection{Uncertainty, surface distance, development cases, and complementarity}
\label{app:stats}
\tblLabel\ref{tab:ci} lists the Dice score of every benchmark (with a 95\% bootstrap interval over cases) and the median HD95
over targets.

\noindent\textbf{Development cases.} The hyperparameters were chosen on the first cases of the seven anatomical
benchmarks. Removing those cases changes the mean over the five benchmarks of \tblLabel\ref{tab:anatomy} and the five
unseen-dataset benchmarks of \tblLabel\ref{tab:main} from 56.3 to 56.1 for \sonnet and from
50.8 to 50.4 for \haiku. \fable was run with the settings chosen for the smaller backbones.

\noindent\textbf{Combining \ours with a trained segmenter.} On identical cases, a per-target oracle choice between
\ours (\sonnet) and VoxTell reaches 71.7 on these ten benchmarks (vs.\ 69.4 for VoxTell
alone) and 67.2 on the four unknown-concept and cross-modality benchmarks of \tblLabel\ref{tab:main} (vs.\ 50.3 for
VoxTell alone). The two methods therefore fail on different targets. A
label-free rule that uses \ours only where the trained model returns an empty or sub-milliliter mask does not capture
this gain (68.8 and 51.5). Those empty masks are mostly on brain metastases, where the MLLM is also weak. Routing by
concept is what works. Using \ours for the concepts outside the trained vocabulary (sarcoma, bladder carcinoma) and the
trained model elsewhere gives 63.8 on these four.

\begin{table*}[t]
\centering
\setlength{\tabcolsep}{4pt}
\begin{adjustbox}{max width=\textwidth}
\begin{tabular}{lr|cccc|ccc}
\toprule
 & & \multicolumn{4}{c|}{Dice [95\% CI]} & \multicolumn{3}{c}{HD95 (mm)} \\
Benchmark & $n$ & \oursf & \ourss & \oursh & VoxTell & \oursf & \ourss & VoxTell \\
\midrule
Pediatric CT & 21 & \na & 62.3 {\scriptsize[60.3, 64.3]} & 56.2 {\scriptsize[53.8, 58.3]} & 72.8 {\scriptsize[69.6, 75.7]} & \na & 12.7 & 7.7 \\
AeroPath & 27 & \na & 84.1 {\scriptsize[83.1, 85.0]} & 84.6 {\scriptsize[83.6, 85.5]} & 89.7 {\scriptsize[88.4, 90.9]} & \na & 38.1 & 14.5 \\
VEELA & 20 & \na & 47.5 {\scriptsize[43.9, 50.6]} & 44.7 {\scriptsize[41.6, 47.5]} & 57.2 {\scriptsize[52.1, 61.4]} & \na & 33.3 & 16.3 \\
HVSMR & 60 & \na & 49.0 {\scriptsize[44.4, 53.6]} & 40.4 {\scriptsize[36.3, 44.4]} & 52.7 {\scriptsize[48.1, 57.3]} & \na & 28.3 & 31.1 \\
ACRIN (PET) & 36 & 56.0 {\scriptsize[47.6, 63.6]} & 34.7 {\scriptsize[27.6, 42.2]} & 24.2 {\scriptsize[17.0, 31.4]} & 62.3 {\scriptsize[51.1, 72.5]} & 53.8 & 359.8 & 32.7 \\
ISBI MS & 21 & 60.8 {\scriptsize[55.1, 66.2]} & 54.7 {\scriptsize[49.6, 59.8]} & 50.8 {\scriptsize[45.9, 55.8]} & 72.4 {\scriptsize[69.1, 75.5]} & 20.2 & 46.6 & 5.1 \\
Adrenal tumor & 52 & 58.3 {\scriptsize[47.1, 69.3]} & 65.9 {\scriptsize[54.7, 75.7]} & 58.0 {\scriptsize[47.7, 67.9]} & 76.7 {\scriptsize[69.5, 83.2]} & 12.5 & 10.1 & 8.6 \\
HCC-TACE & 103 & 59.3 {\scriptsize[53.1, 64.9]} & 51.7 {\scriptsize[45.3, 57.4]} & 38.2 {\scriptsize[31.8, 44.3]} & 66.3 {\scriptsize[60.8, 71.4]} & 17.3 & 25.3 & 13.2 \\
PENGWIN & 100 & \na & 89.8 {\scriptsize[88.7, 90.7]} & 87.0 {\scriptsize[86.1, 87.8]} & 97.6 {\scriptsize[97.4, 97.8]} & \na & 3.1 & 0.8 \\
Brain mets & 105 & 31.6 {\scriptsize[26.0, 37.6]} & 23.5 {\scriptsize[18.3, 28.9]} & 23.5 {\scriptsize[18.7, 28.7]} & 46.7 {\scriptsize[41.0, 52.6]} & 49.1 & 64.8 & 35.8 \\
QIN (PET) & 38 & 60.9 {\scriptsize[54.2, 67.1]} & 54.2 {\scriptsize[45.0, 62.9]} & 42.1 {\scriptsize[30.8, 53.4]} & 69.7 {\scriptsize[60.4, 78.5]} & 41.0 & 43.1 & 10.4 \\
Sarcoma & 102 & 76.5 {\scriptsize[72.6, 80.0]} & 74.4 {\scriptsize[70.4, 78.2]} & 65.9 {\scriptsize[60.3, 71.5]} & 39.3 {\scriptsize[33.0, 45.3]} & 17.0 & 21.3 & 49.2 \\
FedBCa & 47 & 43.0 {\scriptsize[32.5, 53.3]} & 41.8 {\scriptsize[30.5, 52.7]} & 30.8 {\scriptsize[21.1, 40.7]} & 23.3 {\scriptsize[15.9, 31.4]} & 20.9 & 20.5 & 63.0 \\
Esophageal tumor & 1 & 82.2 & 0.0 & 0.0 & 69.1 & 5.0 & 120.1 & 65.2 \\
BraTS & 100 & 77.1 {\scriptsize[74.1, 79.7]} & 73.9 {\scriptsize[70.6, 76.8]} & 62.2 {\scriptsize[59.1, 65.0]} & 88.4 {\scriptsize[86.5, 90.2]} & 3.7 & 4.5 & 2.0 \\
 
\bottomrule
\end{tabular}
\end{adjustbox}
\caption{\textbf{Confidence intervals and boundary error.} We show the Dice scores with 95\% bootstrap intervals over
cases and the median HD95. VoxTell is run by us on the same cases. \oursf was run on the tumor and lesion benchmarks
and BraTS only. The esophageal tumor benchmark is a single volume and therefore has no confidence interval.}
\label{tab:ci}
\end{table*}

\subsection{Backbone study}
\label{app:backends}
\noindent\textbf{Setup.} All backbones run the default pipeline with the same settings on the same ten held-out PENGWIN volumes. We
chose this benchmark because it separates models widely at a cost per volume low enough to evaluate many models. Its three bones
are large, but the sacrum and the left / right distinction are hard. Planning and localization use medium reasoning
effort, and tracing uses the highest effort each model supports. Claude models released before the 5 series receive the
equivalent thinking settings of their API. Gemini models use their native normalized coordinates. Every model was
first checked on synthetic shapes. OpenAI models are called through their provider API, whereas open-weights models
other than Gemma~4 are called through a model router (OpenRouter), whose highest available setting is high reasoning effort. For
these models, we ran the synthetic check with both coordinate formats (pixels and the normalized grid) and kept the
better one. Qwen3.5 (397B and 9B), GLM~4.6V, and Muse Glimmer answer poorly in pixels on large images and therefore use the normalized grid.
All closed models, as well as Qwen3.8 and GLM~5.3 Flash, place boxes with an IoU above 0.75 at the image sizes we render. Qwen3.5
(397B and 9B), GLM~4.6V, and Muse Glimmer fall below that threshold on some image sizes even in their better format, so part of
their low Dice is due to imprecise localization. Gemma~4 accepts at most four
images per request. It therefore receives the same coarse and fine views packed into grid images, and its contour calls hold
two keyframes instead of eight. Claude Opus~5.5, \sonnet, \haiku, and \flash reuse their earlier outputs on these volumes
where the settings match.

\noindent\textbf{Not shown.} Claude Sonnet~4.6 did not finish a volume because 37 of its 93 calls ran out of output
tokens even at a limit of 64K.
Gemma~4 (31B) is not served on the cloud platform we used for Google models.

\begin{figure*}[t]
  \centering
  \includegraphics[width=0.9\textwidth]{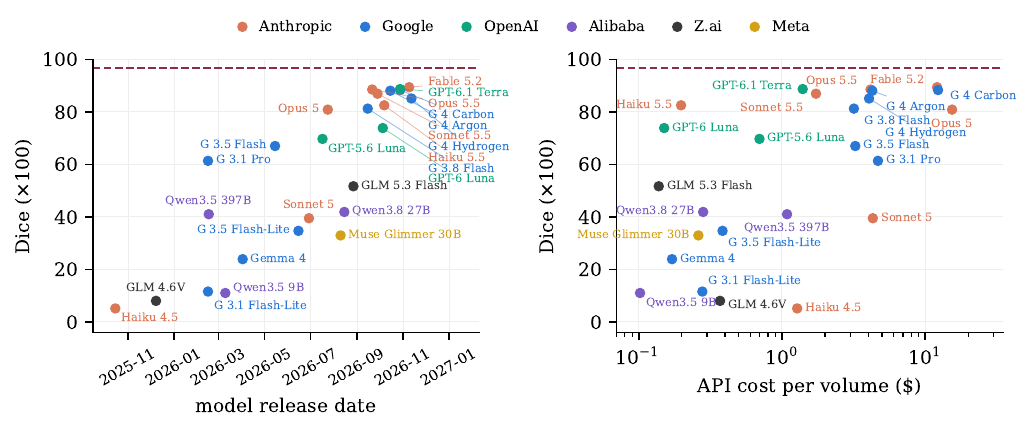}
  \caption{\textbf{Backbones over time and cost.} Dice against release date (left) and cost per volume (right) for the backbones of
  \figLabel\ref{fig:backends}. The dashed line is VoxTell.}
  \label{fig:backend_time}
\end{figure*}

\noindent\textbf{Results.} \tblLabel\ref{tab:backends} lists all 25 finished backbones of \figLabel\ref{fig:backends}
with their release date, Dice, and cost per volume. \figLabel\ref{fig:backend_time} plots Dice against release date
and cost.

\begin{table}[t]
\centering
\setlength{\tabcolsep}{3pt}
\begin{adjustbox}{max width=\columnwidth}
\begin{tabular}{llcrc}
\toprule
Backbone & Released & Dice & 95\% CI & \$ / volume \\
\midrule
Claude Fable 5.2 & 2026-11 & 89.4 & [87.4, 91.2] & 12.12 \\
GPT-6.1 Terra & 2026-10 & 88.7 & [86.7, 90.6] & 1.40 \\
Claude Opus 5.5 & 2026-09 & 88.5 & [86.2, 90.6] & 4.16 \\
Gemini 4 Carbon & 2026-10 & 88.3 & [86.3, 90.2] & 12.31 \\
Gemini 4 Argon & 2026-10 & 88.1 & [85.6, 90.4] & 4.28 \\
Claude Sonnet 5.5 & 2026-09 & 86.9 & [83.4, 89.8] & 1.73 \\
Gemini 4 Hydrogen & 2026-11 & 85.1 & [82.6, 87.6] & 4.06 \\
Claude Haiku 5.5 & 2026-10 & 82.5 & [78.9, 85.6] & 0.20 \\
Gemini 3.8 Flash & 2026-09 & 81.2 & [78.6, 83.9] & 3.18 \\
Claude Opus 5 & 2026-07 & 80.9 & [77.2, 84.1] & 15.44 \\
GPT-6 Luna & 2026-10 & 73.8 & [68.1, 78.9] & 0.15 \\
GPT-5.6 Luna & 2026-07 & 69.7 & [63.8, 75.2] & 0.70 \\
Gemini 3.5 Flash & 2026-05 & 67.0 & [62.9, 70.9] & 3.25 \\
Gemini 3.1 Pro & 2026-02 & 61.4 & [56.4, 65.7] & 4.68 \\
GLM 5.3 Flash$^\dagger$ & 2026-08 & 51.7 & [42.0, 61.2] & 0.14 \\
Qwen3.8 27B$^\dagger$ & 2026-08 & 41.9 & [35.6, 47.9] & 0.28 \\
Qwen3.5 397B$^\dagger$ & 2026-02 & 41.0 & [36.6, 45.4] & 1.09 \\
Claude Sonnet 5 & 2026-06 & 39.5 & [31.3, 47.3] & 4.31 \\
Gemini 3.5 Flash-Lite & 2026-06 & 34.7 & [28.9, 40.1] & 0.38 \\
Muse Glimmer 30B$^\dagger$ & 2026-08 & 33.0 & [28.0, 37.9] & 0.26 \\
Gemma 4 26B$^\dagger$ & 2026-04 & 23.9 & [17.3, 30.3] & 0.17 \\
Gemini 3.1 Flash-Lite & 2026-02 & 11.6 & [8.6, 14.6] & 0.28 \\
Qwen3.5 9B$^\dagger$ & 2026-03 & 11.0 & [5.3, 17.7] & 0.10 \\
GLM 4.6V$^\dagger$ & 2025-12 & 8.0 & [4.7, 12.5] & 0.37 \\
Claude Haiku 4.5 & 2025-10 & 5.2 & [4.0, 6.2] & 1.28 \\
 
\bottomrule
\end{tabular}
\end{adjustbox}
\caption{\textbf{Backbone study on ten held-out PENGWIN volumes.} Cost is the API cost of all calls for a volume at the
providers' list prices, or at the announced introductory price for Gemini~4 Carbon and Hydrogen. $^\dagger$ marks open-weights models.}
\label{tab:backends}
\end{table}

\subsection{Prompt wording}
\label{app:wording}
A trained text-promptable model links words to masks through the vocabulary of its training labels, whereas an MLLM
reads the prompt as language. To test this, we query the same targets with three wordings, namely the dataset
label, a synonym or more specific term, and a plain-language description. For FedBCa, the second wording is ``bladder
cancer'' in place of ``carcinoma'', which also names the organ. For the spleen, the description is ``the fist-sized
organ in the upper left abdomen, behind the stomach and under the diaphragm''. The study covers 12 prompts across five
benchmarks, using held-out cases from PENGWIN, AeroPath, and BTCV, as well as the first 20 cases from sarcoma and FedBCa. We run
\ours (\haiku) and VoxTell on the same cases. \tblLabel\ref{tab:wording} gives the Dice score for every prompt and wording.
\tblLabel\ref{tab:wording_main} of the main paper summarizes the same runs as means over the ten anatomy prompts
(PENGWIN, AeroPath, BTCV) and the two tumor prompts (sarcoma, FedBCa). Each rewording is a single phrasing chosen by
us, so the study measures robustness, not an optimized prompt.

\begin{table}[t]
\centering
\setlength{\tabcolsep}{2.2pt}
\begin{adjustbox}{max width=\columnwidth}
\begin{tabular}{llr|ccc|ccc}
\toprule
 & & & \multicolumn{3}{c|}{\oursh} & \multicolumn{3}{c}{VoxTell} \\
Benchmark & Label & $n$ & label & syn. & descr. & label & syn. & descr. \\
\midrule
PENGWIN & sacrum & 15 & 74.3 & 77.5 & 76.3 & 95.4 & 95.5 & 29.3 \\
PENGWIN & left hipbone & 15 & 87.9 & 88.0 & 86.4 & 97.8 & 97.9 & 97.1 \\
PENGWIN & right hipbone & 15 & 87.1 & 88.1 & 87.5 & 97.4 & 97.4 & 96.9 \\
AeroPath & lung & 12 & 96.0 & 96.1 & 96.0 & 96.9 & 97.0 & 87.6 \\
AeroPath & trachea & 12 & 72.3 & 75.1 & 78.9 & 80.6 & 80.7 & 82.3 \\
BTCV & spleen & 10 & 92.5 & 92.6 & 92.4 & 96.1 & 96.0 & 0.0 \\
BTCV & left kidney & 10 & 89.9 & 90.3 & 86.1 & 94.0 & 91.3 & 93.4 \\
BTCV & gallbladder & 10 & 35.6 & 29.9 & 32.0 & 83.0 & 83.2 & 81.1 \\
BTCV & pancreas & 10 & 61.7 & 63.0 & 61.9 & 84.2 & 84.1 & 0.4 \\
BTCV & aorta & 10 & 85.6 & 79.4 & 84.4 & 91.0 & 89.9 & 90.6 \\
Sarcoma & soft tissue lesion & 20 & 67.7 & 78.7 & 72.2 & 39.2 & 35.6 & 27.1 \\
FedBCa & carcinoma & 20 & 28.2 & 66.8 & 64.5 & 20.2 & 20.3 & 14.1 \\
\midrule
\textbf{Mean} & & & 73.2 & 77.1 & 76.5 & 81.3 & 80.7 & 58.3 \\
 
\bottomrule
\end{tabular}
\end{adjustbox}
\caption{\textbf{Prompt wording.} Dice scores when the same targets are prompted by the dataset label, a synonym, or a plain-language
description.}
\label{tab:wording}
\end{table}

\subsection{A 2D MLLM segmenter applied slice by slice}
\label{app:llmseg_slice}
\begin{figure*}[t]
  \centering
  \includegraphics[width=\textwidth]{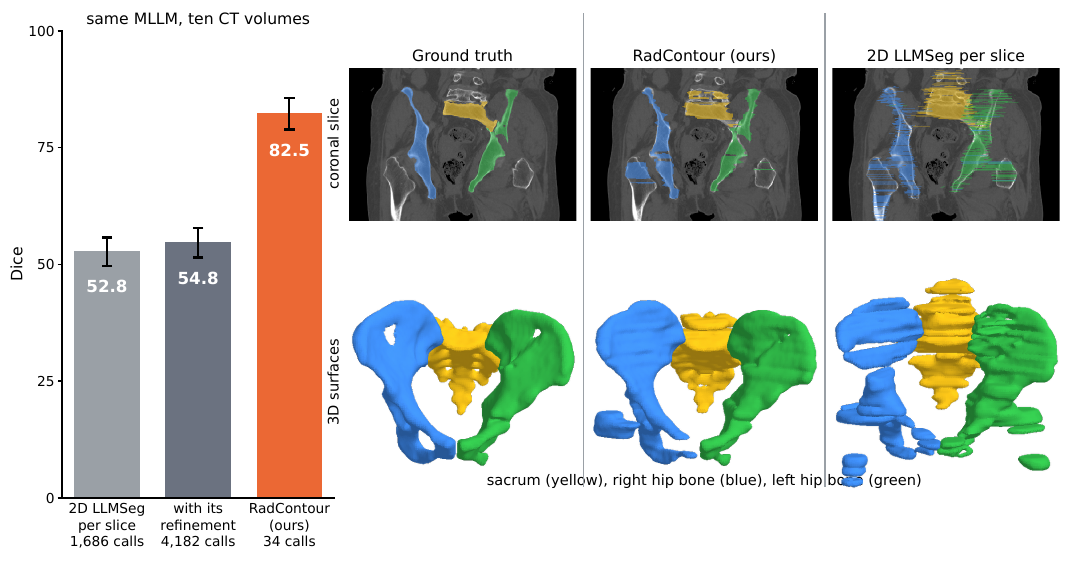}
  \caption{\textbf{LLMSeg on every slice vs.\ \ours, full figure.} \textit{Left:} Dice on the ten held-out PENGWIN
  volumes for LLMSeg applied to every axial slice (without and with its refinement passes) compared with \ours. Both use
  \haiku. Whiskers are 95\% bootstrap intervals over volumes, and the labels give the MLLM calls per volume.
  \textit{Right:} The coronal cut of \figLabel\ref{fig:slice} at its full field of view (top) and the 3D surfaces of
  the same volume seen from the front (bottom). Per-slice outlines do not line up from one slice to the next, so the
  surface of LLMSeg breaks into stacked layers and loose fragments.}
  \label{fig:slice_full}
\end{figure*}

This section gives the setup and the complete numbers of the slice-by-slice experiment of the main paper
(\figLabel\ref{fig:slice} and the paragraph ``Running MLLMs for each slice'' of \secLabel\ref{sec:analysis}).
\tblLabel\ref{tab:llmseg_slice} gives the per-bone results along with precision, recall, HD95, API cost, and calls per volume.
\figLabel\ref{fig:slice_full} shows the Dice scores of the three arms as bars, together with the 3D surfaces of one
volume.
Using \haiku, we run LLMSeg~\cite{hamdi2026llmseg} on every axial slice
of the ten PENGWIN volumes from \secLabel\ref{app:backends} and stack the 2D masks into a volume. LLMSeg is
used unchanged, with its own prompts and code for detection, B-spline segmentation, and refinement. Each slice is
rendered at 1024 pixels in the same CT window as ours. The query is the benchmark prompt plus one sentence stating the
radiological left / right convention. We report two arms that share the detection and segmentation calls, and the
second arm adds LLMSeg's two refinement passes.

\begin{table}[t]
\centering
\setlength{\tabcolsep}{2.2pt}
\begin{adjustbox}{max width=\columnwidth}
\begin{tabular}{lccccccccr}
\toprule
 & Sacr. & L hip & R hip & Dice & Prec. & Rec. & HD95 & \$ & Calls \\
LLMSeg per slice & 45.9 & 55.1 & 57.6 & 52.8 & 41.3 & 77.4 & 70.9 & 0.76 & 1,686 \\
\quad + refinement & 47.7 & 57.9 & 58.9 & 54.8 & 42.9 & 79.6 & 65.5 & 2.06 & 4,182 \\
\midrule
\oursrow \oursh & 72.6 & 87.5 & 87.4 & 82.5 & 77.3 & 90.1 & 5.6 & 0.20 & 34 \\
\oursrow \ourss & 81.4 & 90.0 & 89.4 & 86.9 & 81.3 & 94.6 & 3.4 & 1.73 & 30 \\
\oursrow \oursf & 88.4 & 90.1 & 89.7 & 89.4 & 85.1 & 94.9 & 3.0 & 12.12 & 29 \\
\midrule
VoxTell (trained) & 95.2 & 97.8 & 97.1 & 96.7 & 94.8 & 98.8 & 0.8 & $-$ & $-$ \\
 
\bottomrule
\end{tabular}
\end{adjustbox}
\caption{\textbf{LLMSeg on every slice vs.\ \ours.} Complete results on the ten held-out PENGWIN volumes. LLMSeg uses \haiku.
Precision and recall are voxel-wise, and HD95 is the median in mm. \$ and Calls denote the API cost and the number of
MLLM calls per volume.}
\label{tab:llmseg_slice}
\end{table}

\subsection{Design alternatives}
\label{app:design}
\tblLabel\ref{tab:design} of the main paper changes one component of \ours at a time on anatomy.
\tblLabel\ref{tab:design_all} adds one more row (16 keyframes per contour call). Every row uses \haiku with all other
settings kept at their defaults and is paired with the default configuration on the same volumes. The anatomy set holds 36 held-out
volumes (4 from BTCV, 4 from AMOS, 4 from HVSMR, 10 from PENGWIN, 8 from AeroPath, and 6 from VEELA). None of them were used to
choose a setting. The score is the mean over the six benchmarks, and Calls is the mean number of MLLM calls per volume,
including the plan call.

The rows are defined as follows. \emph{Without the plan}, no plan call is made. Every target receives the soft-tissue window
on CT (or the automatic window otherwise) with no appearance, location, or confounder text, and it is treated as a single solid
structure of 50~mm. \emph{Without fine localization}, the keyframe boxes are interpolated from the boxes of the coarse
call. \emph{Sparser keyframes} uses the base keyframe spacing $\min(8\,\mathrm{mm}, s/2)$, three times the default.
\emph{16 keyframes per call}, \emph{512-pixel crops}, and \emph{lower reasoning effort} change only the contour calls,
replacing the defaults of at most 8 keyframes, 1024 pixels, and the highest effort, respectively. \emph{Without distance blending}, every slice takes the
mask of its nearest keyframe in place of \eqn{eq:sdf}, setting $\widetilde{\mathbf{M}}(\mathbf{u},z)=\mathbf{m}_k(\mathbf{u})$
for the keyframe $z_k$ nearest to $z$. \emph{Without intensity snapping}, the assembled mask is kept as the final output
($\mathbf{M}=\widetilde{\mathbf{M}}$). Both are computed from the default's own outlines, with no new MLLM calls. \emph{MedSAM2 as tracer} prompts MedSAM2~\cite{MedSAM2} with the
default's fine-localization boxes on the same keyframes, so it uses the default's plan and localization calls and no
contour calls.

\begin{table}[t]
\centering
\small
\setlength{\tabcolsep}{4pt}
\begin{adjustbox}{max width=\columnwidth}
\begin{tabular}{@{}l cc@{}}
\toprule
 & \multicolumn{2}{c}{\textbf{Anatomy}} \\
\cmidrule(lr){2-3}
\textbf{Configuration} & Dice $\uparrow$ & calls $\downarrow$ \\ %
\midrule
w/o plan (fixed defaults) & 57.2\,\sd{0.9} & 71 \\ %
\midrule
w/o fine localization & 56.1\,\sd{1.1} & 56 \\ %
\midrule
3$\times$ sparser keyframes & 59.0\,\sd{0.3} & 43 \\ %
16 keyframes per call & 59.3\,\sd{0.6} & 53 \\ %
512-pixel crops & 58.5\,\sd{0.5} & 72 \\ %
Lower reasoning effort & 58.2\,\sd{0.7} & 65 \\ %
\midrule
w/o distance blending & 59.9\,\sd{0.2} & 69 \\ %
w/o intensity snapping & 57.6\,\sd{0.5} & 69 \\ %
\midrule
MedSAM2~\cite{MedSAM2} as tracer & 56.7\,\sd{1.1} & 10 \\ %
\midrule
\oursrow \ours (default) & \textbf{60.0} & 69 \\ %
 
\bottomrule
\end{tabular}
\end{adjustbox}
\caption{\small \textbf{Ablation study, complete.} Mean Dice and mean MLLM calls per volume with \haiku when one
component of \ours is removed or replaced. Rows are grouped by pipeline stage (plan, localization, contouring, and
assembly) and an alternative tracer. All rows are paired on 36 held-out anatomy volumes from six benchmarks. $\pm$ denotes
the standard error of the paired change against the default. Gray marks the default, and bold marks the best value.}
\label{tab:design_all}
\end{table}

\noindent\textbf{Keyframe density.} For \figLabel\ref{fig:density}, each target
of the main runs is re-assembled from every $N$-th keyframe of its own outlines, so no MLLM call is repeated. For $N>1$,
each connected part draws a random starting offset, and we report the mean and the standard deviation over three such
draws. The mask is assembled with the distance blending of \eqn{eq:sdf} and then snapped as in the default. With all
keyframes, the procedure reproduces the main runs exactly. Keyframes denser than the default need new contour calls, so
\figLabel\ref{fig:density} starts at the default.

\subsection{Development analysis}
\label{app:devanalysis}
This section presents the analysis that led to the default, in the order in which it was developed. Unless stated
otherwise, it uses \haiku on 98 held-out cases of seven anatomical benchmarks (the five of \tblLabel\ref{tab:anatomy},
plus BTCV and AMOS), paired with the configuration current at each step. The run-to-run noise of a full replicate on this
set is $\pm0.2$ Dice.

\noindent\textbf{Call load and resolution.} We re-run only the contour calls of 453 development targets with all other
inputs fixed. Against a replicate of the first version's calls (at most 16 keyframes each), putting all of a target's
keyframes in one call changes 3D Dice by $-$0.4. Limiting calls to at most 12, 8, 4, and 1 keyframes changes it by +0.4,
+0.9, +1.0, and +1.3, respectively. On targets with more than 8 keyframes, the latter changes are +1.7, +2.1, +2.9, and
+3.7. Crops of 768, 1024, and 1536 pixels on the long side beat 512-pixel crops by +1.0, +1.8, and +0.7, respectively
(s.e.\ 0.3 to 0.4).

\begin{table}[t]
\centering
\setlength{\tabcolsep}{5pt}
\begin{adjustbox}{max width=\columnwidth}
\begin{tabular}{ll|c}
\toprule
Boxes from & Outlines from & Anatomy \\ %
\midrule
\sonnet & MedSAM2~\cite{MedSAM2} & 62.8 \\ %
\oursrow \sonnet & MLLM tracing (ours) & 66.5 \\ %
 
\bottomrule
\end{tabular}
\end{adjustbox}
\caption{\textbf{Is a trained segmenter needed?} MedSAM2 replaces our MLLM tracer and is prompted by the boxes of our
fine localization on the same keyframes. Mean Dice over the five anatomical benchmarks of \tblLabel\ref{tab:anatomy}.}
\label{tab:tracer}
\end{table}

\noindent\textbf{Recognition and tracing.} We break down the Dice lost by the default pipeline on the held-out set by
error type on each axial slice. Of the lost Dice, 20\% comes from imperfect outlines inside a correct box, 32\% from
slices where the mask is off the target or missing, 26\% from boxes that lie partly off the target, and 22\% from masks
on slices without the target. Two oracles compare the MLLM's localization with ground-truth localization. A trained
segmenter, MedSAM2~\cite{MedSAM2}, reaches 82.0 Dice from ground-truth boxes and 48.7 from the MLLM's boxes on the
first 20 cases of every benchmark. Ground-truth outlines at the keyframes of an earlier configuration (three times
denser keyframes than the first version, at most 8 per contour call) give 91.8 instead of 56.8.
\tblLabel\ref{tab:tracer} shows the effect of replacing the MLLM tracer with
MedSAM2, prompted by the boxes of our fine localization on the same keyframes.

\subsection{Target size and the gap to a trained segmenter}
\label{app:sizegap}
\begin{figure}[t]
  \centering
  \includegraphics[width=\linewidth]{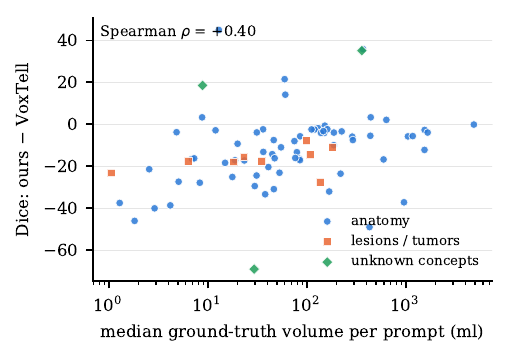}
  \vspace{-14pt}
  \caption{\textbf{The gap to a trained segmenter depends on target size.} Each point is one of the 79 prompts of our
  17 benchmarks. It shows the Dice of \ours (\sonnet) minus that of VoxTell on identical cases, against the
  median ground-truth volume of the prompt. Below 5~ml, \ours trails by 30 points, and above 50~ml it trails by 7 to 9. On the unknown concepts, it
  leads.}
  \label{fig:sizegap}
  \vspace{-6pt}
\end{figure}

For each of the 79 prompts, \figLabel\ref{fig:sizegap} plots the Dice difference between \ours (\sonnet) and the best
trained model on identical cases against the prompt's median ground-truth volume.

\subsection{Optional variants}
\label{app:variants}
Two additions improve specific regimes. Neither is part of the reported default, so the main results use one
configuration with the MLLM as the only vision component.

\noindent\textbf{In-context examples.} When a concept is ambiguous or defined by a labeling convention, one to three
labeled volumes of \emph{other} patients can be shown to the model as examples. Their masks are outlined in a
distinct color in the plan, localization, and contour calls, and nothing is trained. With the first pipeline version
and \haiku on 20 cases each from ten benchmarks, one example adds +3.2 Dice (s.e.\ 2.1), and three examples add +4.7
(s.e.\ 2.5), with gains on 7 of 10 benchmarks. Bladder carcinoma rises from 26.5 to 41.3 with one example and to 49.9
with three. With one example, sarcoma rises from 70.7 to 81.0, and breast cancer in PET rises from 47.1 to 57.6. Examples
help where the \emph{concept} is unclear, not where perception is the limit (small metastases, subtle liver lesions).
With a single example, the model can anchor on the example's side or size, so the example text omits both. No trained
text-promptable baseline can use such examples.

\noindent\textbf{A second scan for small lesions.} After the first pass, one extra scan of whole slices every 5~mm asks
for small spots only. New parts under 1~ml are kept and traced like first-pass parts. On all 105 brain-metastasis
volumes, with \haiku and otherwise default settings, Dice rises from 23.4 to 28.2 (+4.8, s.e.\ 1.2), with 48 cases
better and 12 worse. On the 65 cases not used to inform any design choice, the gain is +2.9. Lesion recall rises from 6.9\% to
20.5\%, for about \$0.05 per volume. On the seven anatomical benchmarks, the scan changes Dice by only +0.09
(s.e.\ 0.03), since it is triggered only for multi-lesion targets.

\section{Cost Trends}
\label{app:cost}
\begin{figure*}[t]
  \centering
  \includegraphics[width=\textwidth]{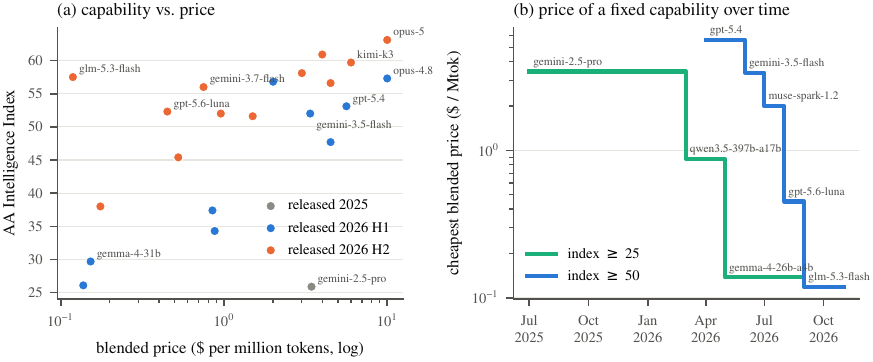}
  \caption{\textbf{The price of a given MLLM capability falls quickly.} (a) Capability (Artificial Analysis
  Intelligence Index~\cite{artificialanalysis2026}) against blended API list price for the 22 indexed models in our price
  sheet, by release period. (b) The cheapest model available at each date whose index reaches 25 or 50. The blended
  price is (3 input + 1 output) / 4, in \$ per million tokens. Claude Haiku~5.5 and Sonnet~5.5 have no index in the sheet
  and are not shown.}
  \label{fig:pricetrend}
\end{figure*}

\ours is billed per volume in tokens, so its cost follows the price of MLLM capability rather than that of data and
training. \figLabel\ref{fig:pricetrend} shows how the price of a fixed capability changes over time. These
are API providers' list prices and do not measure the providers' own compute cost. Moreover, the index is a general
benchmark, not a medical one.

\clearpage
\section{Prompt Templates}
\label{app:prompts}
The four zero-shot prompts are given verbatim. Text in angle brackets is filled in per call with the target, the
modality, the plan's description of the target, the view counts, and the display convention. The prompts write a
pixel as [x, y], which is $\mathbf{u}=(u,v)$ in the notation of the Method. A box [x1, y1, x2, y2] in view pixels is
mapped back to the shown slice, and the boxes of the fine call are the boxes $\mathbf{B}_z$. For backbones that answer on a normalized grid, these
coordinate instructions and JSON formats are rewritten for that grid (\secLabel\ref{app:method}).

\begin{promptbox}{Plan}
\begin{lstlisting}[style=promptstyle]
You are an expert radiologist preparing to segment structures in a 3D <modality> volume <of the body region>.
<overview views> For each target below give (interpret every target in the context of THIS scan's body region and modality):
- "window": <choice of a display window, CT only>
- "appearance": one sentence on how the target looks on this modality (intensity vs. surroundings, texture, shape).
- "location": one sentence on where it lies (landmarks, side of the body).
- "multiplicity": "single" if it is one connected structure, "multiple" if it can appear as several separate pieces or lesions.
- "shape": "solid" if a cross-section is a filled region, "shell" if a cross-section is a thin ring or band around something else (e.g. a wall or a surface layer), "tubular" if it is a vessel / duct / canal whose cross-section is a small disc, "branching" for a branching tree of thin vessels or airways.
- "size_mm": its typical largest diameter in mm (a number).
- "confounders": one sentence naming the nearby structures or normal findings that look similar on this modality and are NOT the target (e.g. normal physiological uptake, adjacent organs or vessels), and how to tell them apart.

Targets:
<numbered list of target prompts>

Return ONLY JSON: {"targets": [{"id": 1, "window": "...", "appearance": "...", "location": "...", "multiplicity": "single", "shape": "solid", "size_mm": 50, "confounders": "..."}, ...]}
\end{lstlisting}
\end{promptbox}

\begin{promptbox}{Coarse localization}
\begin{lstlisting}[style=promptstyle]
You are an expert radiologist locating targets in a 3D <modality> scan.
You are shown <number of views> images of ONE scan. VIEW 1..<N> are axial slices ordered from inferior (feet) to superior (head); <one coronal and one sagittal context view follow>
Display conventions: each axial view is an <axial display convention>. The coronal view is a <coronal display convention>. The sagittal view is a <sagittal display convention>.
Each caption states the view's size in pixels and its position in the volume.

Targets:
<numbered list of target prompts>

For EVERY target and EVERY view in which the target is visible, give a tight bounding box [x1, y1, x2, y2] in the <image pixel coordinates>. Box the target in every view where any part of it appears, including where it is small. If the target has several separate parts or lesions in one view, give one box per part. Leave out views where the target is not visible; if a target is not visible anywhere, give an empty list. Do not confuse left and right: use the display convention above.

Return ONLY JSON: {"targets": [{"id": "T1", "detections": [{"view": 3, "box": [x1, y1, x2, y2]}, ...]}, ...]}
\end{lstlisting}
\end{promptbox}

\begin{promptbox}{Fine localization}
\begin{lstlisting}[style=promptstyle]
You are an expert radiologist locating "<prompt>" in a 3D <modality> scan.
About the target: <appearance, location and confounders from the plan>
You are shown <number of views> axial slices (VIEW 1..<number of views>) of ONE scan, ordered from inferior (feet) to superior (head). Each view is an <axial display convention>. Each caption states the view's size in pixels and its slice position.

For EVERY view in which any part of "<prompt>" is visible, give a tight bounding box [x1, y1, x2, y2] around it in the <image pixel coordinates>. If it has several separate parts in one view, give one box per part. Leave out views where it is not visible. Do not confuse left and right: use the display convention above. Do not box neighbouring structures.

Return ONLY JSON: {"detections": [{"view": 1, "box": [x1, y1, x2, y2]}, ...]}
\end{lstlisting}
\end{promptbox}

\begin{promptbox}{Keyframe contouring}
\begin{lstlisting}[style=promptstyle]
You are an expert radiologist delineating "<prompt>" in a 3D <modality> scan.
About the target: <appearance, location and confounders from the plan> <shape note>
You are shown <K> keyframes of axial slices, ordered from inferior to superior. Keyframe k is a pair of images: FULL k is the whole axial slice (small) with a yellow rectangle, and CROP k is the zoomed region inside that rectangle. Every slice is an <axial display convention>. Use FULL k to recognise the anatomy and decide whether "<prompt>" is present in that slice; trace it in CROP k. The target's shape and size change from slice to slice, so judge each slice on its own anatomy.

For EACH keyframe:
- if "<prompt>" is present in the slice, give one closed loop of 8-40 points [x, y] lying ON its boundary, in order around it, in CROP k pixel coordinates (origin top-left, x right, y down; sizes are in the captions). A smooth closed curve is drawn through the points, so use more points where the boundary bends.
- if it has several separate parts in that slice, give one loop per part; if it has an interior hole (a ring), add a loop with "role": "hole".
- if it is NOT present in that slice, set "present": false and give no loops.
Trace only "<prompt>", not neighbouring structures.

Return ONLY JSON: {"crops": [{"crop": 1, "present": true, "loops": [{"role": "outer", "points": [[x, y], ...]}]}, ...]}
\end{lstlisting}
\end{promptbox}

\end{document}